\documentclass{article}

\PassOptionsToPackage{numbers, sort&compress}{natbib}

\usepackage[final]{neurips_2024}

\usepackage[utf8]{inputenc} 
\usepackage[T1]{fontenc}    
\usepackage{hyperref}       
\usepackage{url}            
\usepackage{booktabs}       
\usepackage{amsfonts}       
\usepackage{nicefrac}       
\usepackage{microtype}      
\usepackage{xcolor}         
\usepackage{graphicx}
\usepackage{subcaption}
\usepackage{booktabs}
\usepackage{listings}
\usepackage{xspace}
\usepackage{spverbatim}

\usepackage[accsupp]{axessibility}  

\newcommand{\methodname}{FlexCap\xspace}
\title{\Large{\methodname: Describe Anything in Images in Controllable Detail}}

\author{%
  Debidatta Dwibedi \\
  Google Deepmind \\
  \texttt{debidatta@google.com} \\
   \And
   Vidhi Jain\thanks{Work done as a student researcher at Google Deepmind.} \\
   Carnegie Mellon University \\
   \texttt{vidhij@andrew.cmu.edu} \\
   \AND
   Jonathan Tompson \\
   Google Deepmind \\
   \texttt{tompson@google.com} \\
   \And
   Andrew Zisserman \\
   Google Deepmind \\
   \texttt{zisserman@google.com} \\
   \And
   Yusuf Aytar \\
   Google Deepmind \\
   \texttt{yusufaytar@google.com} \\
}

\begin{document}

\maketitle

\begin{abstract}
We introduce \methodname, a vision-language model that generates region-specific descriptions of varying lengths. \methodname is trained to produce length-conditioned captions for input boxes, enabling control over information density, with descriptions ranging from concise object labels to detailed captions. To achieve this, we create large-scale training datasets of image region descriptions with varying lengths from captioned web images. We demonstrate FlexCap's effectiveness in several applications: first, it achieves strong performance in dense captioning tasks on the Visual Genome dataset. Second, we show how FlexCap's localized descriptions can serve as input to a large language model to create a visual question answering (VQA) system, achieving state-of-the-art zero-shot performance on multiple VQA benchmarks. Our experiments illustrate FlexCap's utility for tasks including image labeling, object attribute recognition, and visual dialog. Project webpage: \url{https://flex-cap.github.io}.
\end{abstract}

\section{Introduction}
How does one describe the world around us, not just in broad strokes but with the ability to zoom in and out, capturing both the grand scene and the minute details? Imagine pointing at a bustling market scene and asking, "What's happening here?" and receiving a vivid description, not just of the market as a whole, but also a detailed account of the interactions between vendors and customers, the vibrant colors of the goods on display, or even a specific item held by a passerby. This ability to controllably focus and describe visual content is what we call \textit{flexible captioning}.

\begin{figure*}[t]

\begin{center}
  \includegraphics[width=\linewidth]{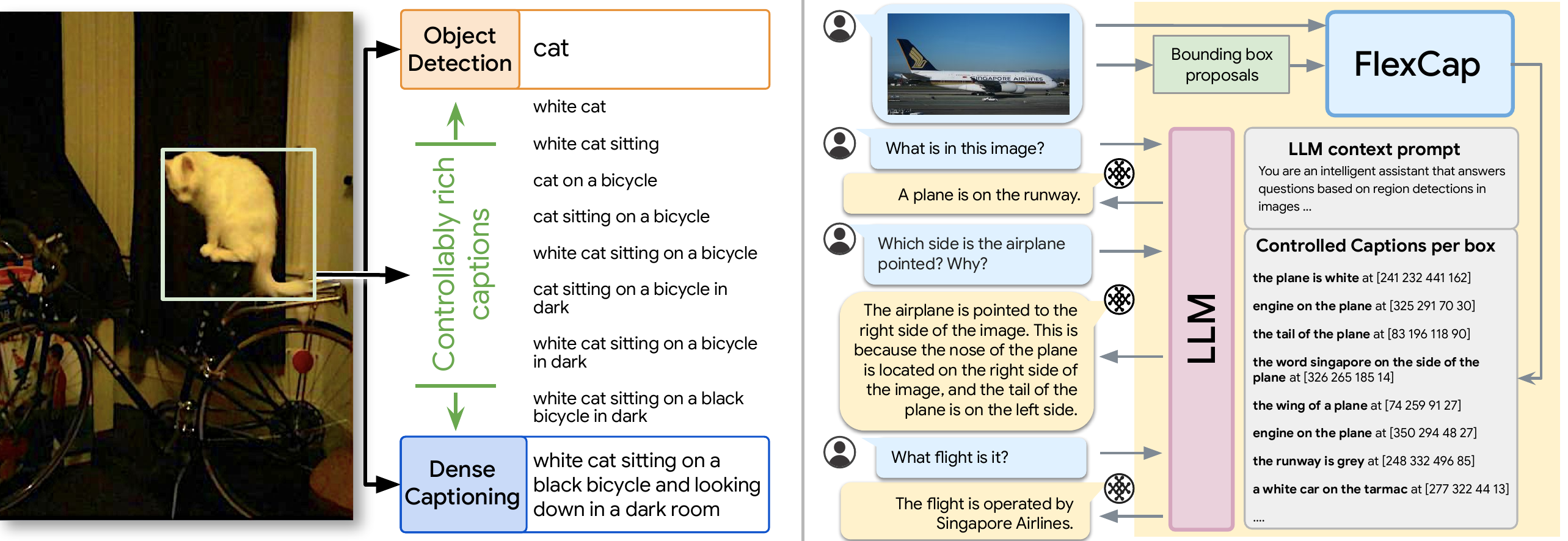}
\end{center}
\caption{ \textbf{\methodname} generates controllably rich localized descriptions for any region in an image as shown on the left. It has the flexibility to produce captions in a controllable manner which allows the full spectrum of valid descriptions  to be explored from short object category names to fully-detailed captions. On the right, we demonstrate that rich localized captions generated by \methodname, when coupled with large language models (LLMs), enable zero-shot visual question answering. 
}
\label{fig:teaser}
\vspace{-.15in}
\end{figure*}

Traditional image captioning models, while adept at capturing the gist of an image, often struggle to pinpoint specific objects or attributes. On the other hand, object detection systems excel at localizing elements but may lack the vocabulary to describe them comprehensively. Dense captioning~\cite{johnson2016densecap} attempts to bridge this gap by generating captions for multiple regions, but its expressiveness is limited by existing datasets.
In this work, we introduce a model called \methodname that bridges the gap between holistic image understanding and localized inquiry. This enables the generation of captions that are both spatially precise and semantically rich (see Fig.~\ref{fig:teaser} (left)) by specifying a region of interest and the desired level of detail in terms of number of words in the predicted caption. This allows us to integrate the strengths of image captioning, object detection, and dense captioning into one model.

To be able to train such a model, we require a dataset of images where many boxes are labeled with short and long descriptions. We propose a method to generate  triplets of (i) image, (ii) a proposed region within the image, and (iii) a caption of a particular length, by using an open-vocabulary object detector to label regions from captions of an image-text pair dataset. We demonstrate this at two scales: 200 million triplets using YFCC100M~\cite{thommeeyfcc100m} captioned images; and 32 billion triplets using the WebLI~\cite{chen2022pali} captioned dataset.
Training \methodname on these datasets enables the model to generate spatially and semantically rich representations (bounding boxes and their descriptions) that focuses on  objects, their attributes, and their contextually changing descriptions. 

We evaluate the effectiveness of the descriptions generated with \methodname by representing images in detail by describing its constituent objects and regions. A popular and effective strategy is to directly provide visual features as input tokens to LLMs ~\cite{Alayrac2022FlamingoAV,li2022blip,Li2023-px,driess2023palm, chen2022pali,liu2023llava,you2023ferret}. Instead we directly provide textual descriptions of images as input to the LLM to evaluate \methodname similar to~\cite{yang2022empirical,hu2023promptcap,shao2023prompting}. We show that this human interpretable representation, when combined with the power of LLMs, enables visual question answering and dialog. An example is shown in  Figure~\ref{fig:teaser}(right). We also demonstrate that this combination can result in performance that is competitive with state-of-the-art VLM models on zero-shot image and video question answering benchmarks. 

Our key technical contributions are:
(i) controllable localized visual descriptions, using word count as a proxy for complexity to modulate the output of a generative language model
(ii)  a large-scale dataset generated for image-text-box triplets to enable training of our model; 
(iii) the human-interpretable representations produced by FlexCap with the help of LLMs is comparable or exceeds performance of state-of-the-art VQA methods;
and (iv) demonstrating that our dense captioning methods outperform baselines under comparable scenarios.
\vspace{-1em}
\section{\methodname}
\label{sec:arch}

\begin{figure*}[t]
\begin{center}
   \includegraphics[width=\linewidth]{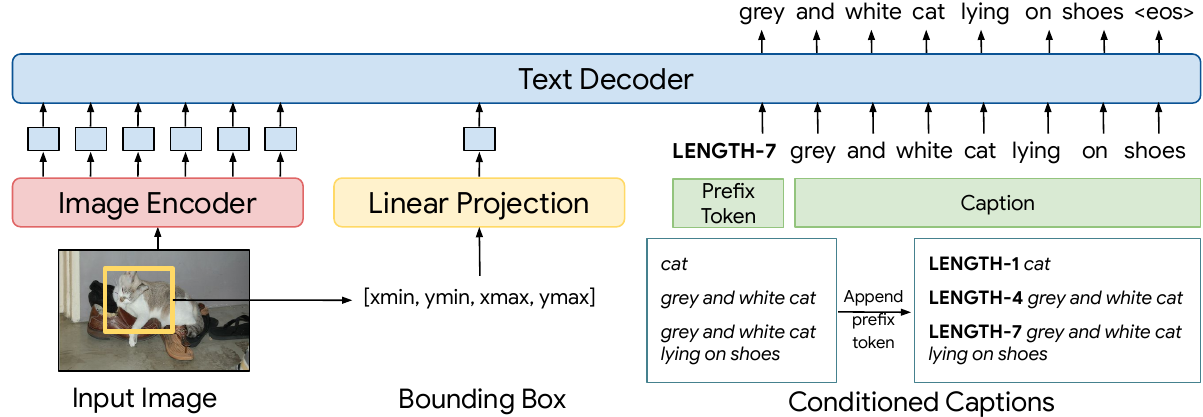}
\end{center}
   \caption{\textbf{Architecture and Training Setup.} \methodname outputs a length-controlled caption of the object contained in the bounding box by taking (left) an image, (middle) coordinates of a bounding box  and (right) the length prefix and caption, as inputs. The training loss is the standard next-word prediction loss that is used to train image captioning models.}
\label{fig:architecture}
\vspace{-.25in}
\end{figure*}

\noindent\textbf{Length conditioning.}
For the same region in an image, there may be multiple valid captions. In the input image shown in Figure~\ref{fig:architecture}, all the following descriptions are correct: \textit{cat}, \textit{grey and white cat}, \textit{grey and white cat lying on shoes}. 
Clearly, the task of describing a bounding box in the image does not have only one right answer.  
We equip the model with the capability of producing outputs of a desired length by utilizing the idea of \textit{length} conditioning. We condition the input by simply using an additional token indicating the desired length of the output caption.
Training with length-conditioning is useful for three key reasons.
First, the number of words used to describe is often proportional to the information content. We train the model to predict the next word in the sequence while accounting for the desired length, thereby the model learns to modulate the amount of information in the generated text.
Second, length conditioning allows users to control the output of the model, further enabling the use of a single model for many diverse tasks.
Third, the length prefix provides a better conditioned initial state for the captioner.
Figure~\ref{fig:dataset_generation} shows how the same box might have more than one ground truth caption \verb|<s> a dog <e>| or \verb|<s> dog playing with a frisbee <e>|. If we use the first caption as ground truth and the words \verb|<s> a dog| as the seen text, the next-word prediction loss encourages the model to increase its score for the \verb|<e>| token and decrease the score for the word \verb|playing| due to the softmax loss. 
Whether this occurrence will be a problem depends on the dataset statistics. To quantify the prevalence of this problem, we compute a statistic: for each image box, we consider all pairs of captions and measure the fraction of pairs sharing prefix words. For instance, consider a box that has three captions: \verb|<s> dog <e>|, \verb|<s> dog playing <e>|, and \verb|<s> dog playing with a frisbee <e>| which share a prefix  \verb|<s> dog|. While another caption for the same box, \verb|<s> brown dog <e>|, does not share a prefix with captions beginning with \verb|<s> dog <e>|. After averaging this metric across all images in our Localized Captions Dataset from WebLI (introduced later in Section~\ref{sec:data}), we found that 30.8\% of all caption pairs share a prefix. Using a length conditioning token instead of \verb|<s>|, the probability of prefix matching decreases from 30.8\% to 11.1\%. The length conditioning helps the model in distinguishing between captions with the same prefix while also providing the model with a novel capability during inference.

\noindent \textbf{Architecture.} Our objective is to train a model that takes as input an image and a region of interest and outputs a description {\em of a desired length} of the region spanned by the box. We present \methodname's architecture in Figure~\ref{fig:architecture}. The model takes an image, the coordinates of a bounding box and the conditioning tokens as input, and outputs a textual description of visual contents within the specified bounding box. Our model mainly consists of an image encoder (i.e.\ SOViT-400M/14~\cite{alabdulmohsin2023getting}) and a transformer-based text-decoder. We pass the image through the vision model to produce outputs of dimensions $n \times d$ (where $n$ is the number of patches and $d$ is the embedding size). We pass the bounding box coordinates (of dimensions $1\times4$) through a linear layer to produce the coordinate features (of dimension $1 \times d$). Both vision features and normalized bounding box  features are concatenated to form input of dimension $(n+1) \times d$ to a text decoder. The text decoder consists of a stack of $L$ Transformer layers. We use a decoder-only architecture in which all the vision and bounding box tokens remain unmasked but the text tokens are masked in a causal manner to enable next-word prediction training. We add all the vision tokens and bounding box coordinate tokens so that the text decoder can access all of the visual context in the image and the exact bounding box location.
In this work, we train a text decoder composed of 12 self-attention transformer layers with a dimensionality of 768 and 12 attention heads. 

In total, \methodname has 590M parameters with 428M comprised of the image encoder (SOViT) and the remaining parameters in the text decoder.
A linear layer transforms the 1152-dimensional output from the vision encoder into a 768-dimensional input for the text decoder. We initialize the vision encoder with SigLIP~\cite{zhai2023sigmoid} weights, which is a contrastively trained image encoder using web-scale vision-text pairs from WebLI. We do not freeze the vision encoder during training.

\noindent \textbf{Loss.} We train \methodname to predict the next token of the text. The text tokens are prefixed with the desired length of the caption and appended with an end of sentence token \verb|<e>| to indicate the end of the caption. The target text tokens are obtained by shifting the padded text by 1. This is common training methodology for training generative language models like GPT~\cite{brown2020language} or SimVLM~\cite{wang2021simvlm}. The loss is a classification loss over all the words present in the vocabulary. The loss is ignored over the padded tokens that are used to keep the size of the outputs same for all the captions in the batch. Formally, we represent a data sample as a triplet $ T=(X, B, W)$ consisting of image $X$, bounding box $B$ and captions $W$, where $W=\{\text{LENGTH-K}, w_1,w_2,...w_k\}$. To enable batch training, we pad the tokenized captions to a fixed size $M$. For a given data triplet, our objective is to maximize the following log-likelihood.
$l(X, B, W) =\sum^M_{i=1} \log p(w_i| w_{<i}, X, B)$.
 Assume that we have a dataset $\mathcal{D} = \{T_1, T_2, ..., T_N\}$. The overall loss function is:
$$ L(D) = \sum_{j=1}^N l(X_j , B_j, W_j) =  \sum_{j=1}^N \sum^M_{i=1} \log p((w_j)_{i}| (w_j)_{<i}, X_j, B_j)$$ 

\noindent \textbf{Implementation.} We implement this model using the JAX framework~\cite{jax2018github}. We train the entire model for about $400K$ steps using the AdamW optimizer with a cosine learning rate schedule. The maximum learning rate is $1.6 \times 10^{-4}$ with 10K warm-up steps. We use a weight decay of $0.05$. We train with a batch size of $4096$ and image resolution of $224 \times 224$. We use a maximum text sequence length of $32$ tokens. For each image in the batch, we sample a maximum of 8 bounding boxes.

\noindent \textbf{Inference.} At inference time, we provide an image, the target bounding box, and the desired length  as input.  We then decode in an auto-regressive manner till the end of caption token \verb|<e>| is encountered or the maximum number of decoding steps is reached.  We use greedy decoding in all experiments unless otherwise stated. However, we can use standard sampling techniques used in text-generation like beam search, temperature sampling, or nucleus sampling~\cite{holtzman2019curious} to generate multiple captions.

\section{Localized Captions Dataset}
\label{sec:data}
\begin{figure}[t]
\begin{center}
   \includegraphics[width=\linewidth]{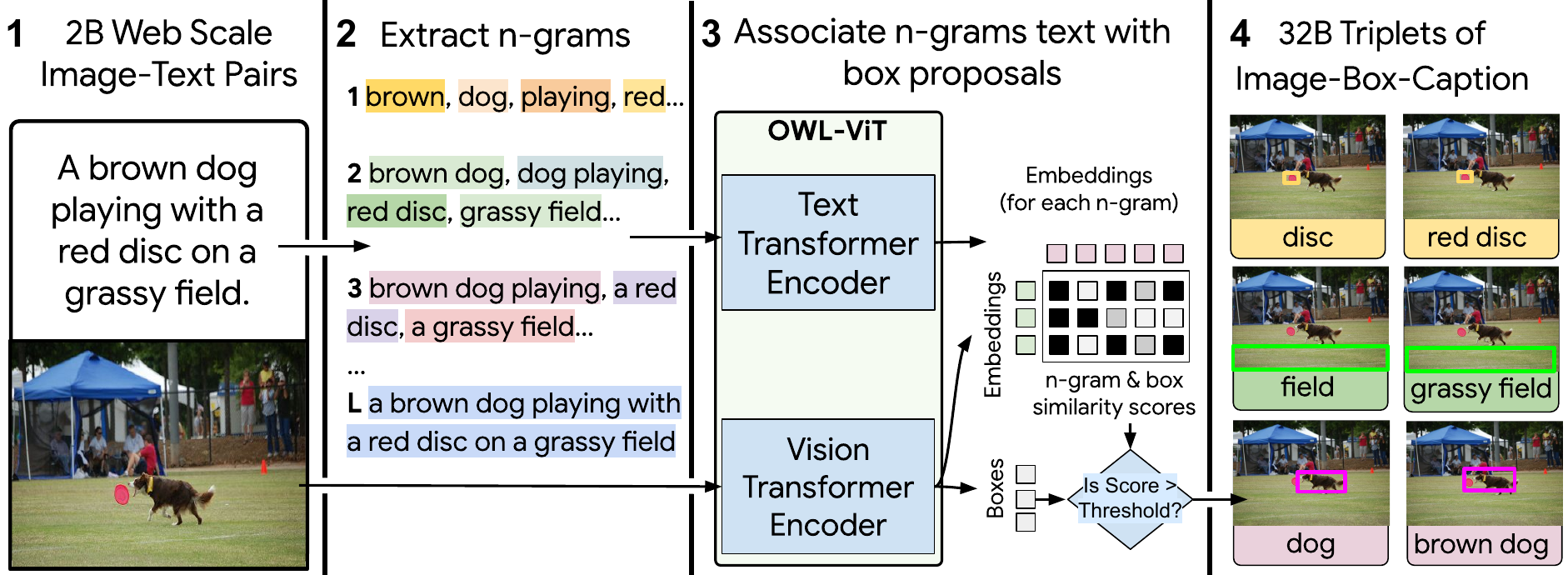}
\end{center}
   \caption{\textbf{Dataset Generation.} We use OWL-ViT to generate a dataset of triplets of image, bounding box and captions from a web-scale dataset of noisy image-text pairs. Increasing levels of richness in captions is captured through different length descriptions for each box.}
\label{fig:dataset_generation}
\vspace{-.20in}
\end{figure}

In order to train the \methodname model, we build a large scale dataset of image region descriptions of varying lengths. In the following section we describe how we produce such a dataset from existing image-text paired datasets. We leverage the web-based image-caption pairs datasets (like  WebLI~\cite{chen2022pali} and YFCC100M~\cite{thommeeyfcc100m}) to create a localized captions dataset. 
The dataset generation pipeline is shown in Figure~\ref{fig:dataset_generation}. First, we create \textit{text queries} using n-grams from the caption of the image: e.g. ``dog",``brown dog", ``brown dog playing with a disc". We specifically create n-grams where $n=\{1,2,\cdots, 8\}$ and then filter out incomplete captions like ``with a red", ``dog playing with". More details about the filtering step are mentioned in the appendix. 
Then we use the filtered n-grams as  \textit{text queries} for pre-trained region proposal models (i.e.\  OWL-ViT~\cite{minderer}) to extract boxes and select text-box pairs based on the similarity score ($>$ 0.1). Multiple n-grams may match for a box, and this results in several ways of describing a box in the image as shown in Col. 4 in Figure~\ref{fig:dataset_generation}. 

\noindent\textbf{WebLI.} This data collection technique on the WebLI dataset results in 32 billion image-box-caption triplets from 2 billion images without requiring new manual annotation. Our captions show a rich vocabulary that is close to common language used to describe objects in the context of an image. If we use MS-COCO's vocabulary then all humans in the dataset would get labeled as \textit{person}. However by building our vocabulary in a bottom-up manner we end up with captions that contain more informative words such as \textit{baby}, \textit{nurse}, \textit{policeman}, \textit{firefighter}, or \textit{baseball player} to describe the \textit{person} class. Please refer to the appendix for details of dataset statistics and examples.

\noindent\textbf{YFCC100M.} We also create a localized captions dataset using YFCC100M images. Specifically we use the same 14M images as the CLIP paper. The dataset creation method results in $\sim11$M images with at least one valid box. On average each image has $20$ boxes, making the size of this dataset $\sim0.2$B image-box-caption triplets. The number of YFCC100M triplets is ~160 times smaller than the localized captions dataset created from WebLI.

As both OWL-ViT and the CLIP subset of YFCC100M are publicly available, the resulting localized captions dataset can be generated with open-source models and public datasets. Since the WebLI dataset is not publicly available yet, YFCC100M triplet-dataset can  serve as a reproducible benchmark. Concurrently large-scale grounded image-text dataset generation pipelines have also been proposed in \cite{peng2024grounding} and ~\cite{wang2023allseeing}.

\vspace{-1em}
\section{Experiments}
\begin{figure}[t]
\begin{center}
   \includegraphics[width=0.9\linewidth]{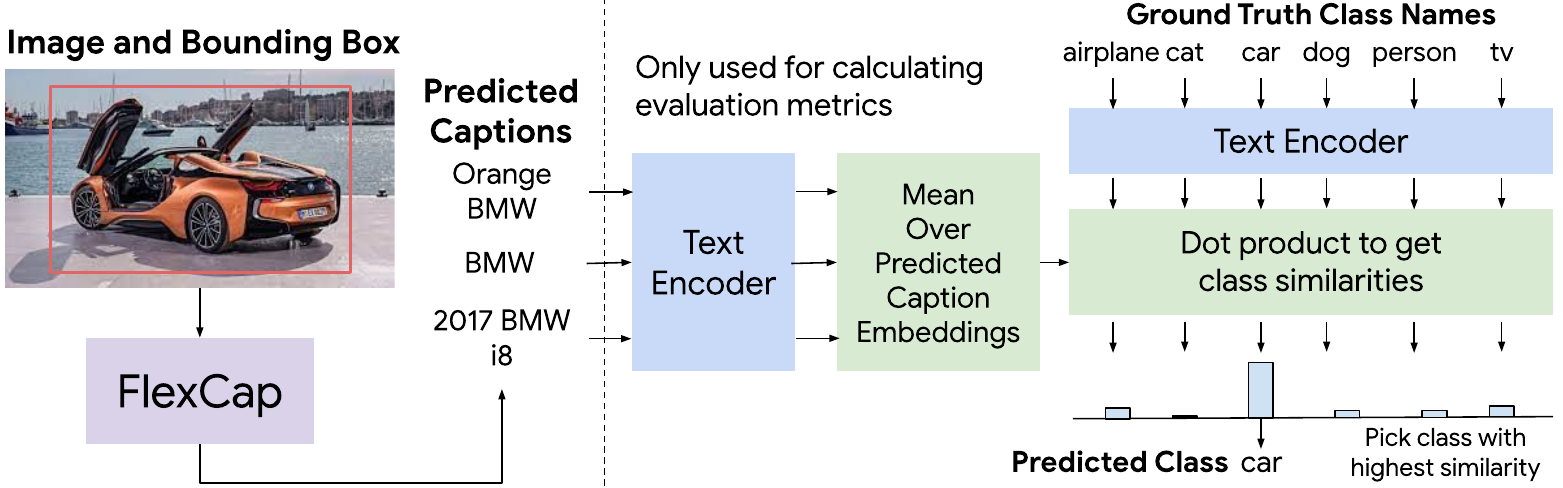}
\end{center}
   \caption{\textbf{Evaluating open-vocabulary outputs} from FlexCap using the CLIP~\cite{radford2021learning} text encoder.
   }
\label{fig:evaluation}
\vspace{-.15in}
\end{figure}

\begin{figure}[t]
\begin{center}
  \includegraphics[width=.9\linewidth]{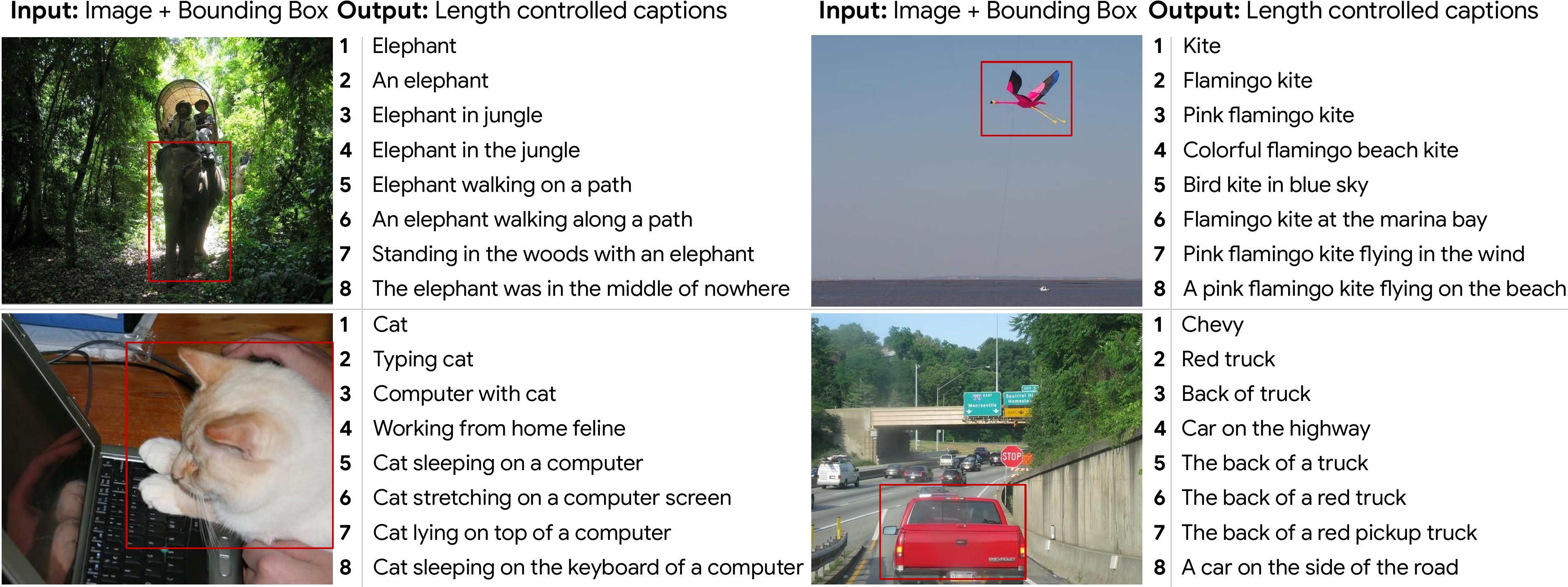}
\end{center}
  \caption{\textbf{Examples of length controlled captions generated by \methodname.} Note that attributes (``pink flamingo kite") and context (``in the jungle") are generated as the length increases.}
\label{fig:length_control}
 \vspace{-.15in}
\end{figure}

\label{sec:fidelity}
\subsection{Correctness and Compliance of Generated Captions}

\textbf{Correctness.} In this experiment, we  solely evaluate the recognition capabilities of our model in a zero-shot manner.  We use the region classification task~\cite{wu2024clim,zhong2022regionclip} on the MS-COCO dataset to assess how well our model recognizes objects at different scales and under occlusion. In this task, the image and the ground truth bounding box are provided as input to the model such that it produces a short description of what is contained in the bounding box. 
Our region classification pipeline (Figure~\ref{fig:evaluation}) passes the input image (of size $448\times448$) through \methodname which generates 20 captions each for 4 caption lengths (1, 2, 3, 4) via nucleus sampling. These captions are then mapped to object class names using CLIP's~\cite{radford2021learning} text encoder. By comparing the mean of the text embeddings of predicted captions with those of ground truth class names, we obtain the classification scores.
We report the results in Table~\ref{tab:region_classification}.
We find \methodname outperforms contrastively trained approaches used for region classification. Furthermore, we observe a significant boost (~13\% mAP) in performance by producing multiple captions for each box. One reason for the boost in performance may be directly comparing text embeddings to produce classification scores as compared to baselines which use dot product of image and text embeddings.  

\textbf{Compliance.} In Figure~\ref{fig:length_control}, we show qualitative examples of the \methodname model producing different length captions for the same box. Note how the model progressively adds more information about the object by incorporating context in the longer sentences (\textit{in the jungle}), attributes (\textit{pink flamingo kite}), and alternative nouns (\textit{chevy}, \textit{feline}). We also measure how well our model complies to the desired caption length. To do so, we take 1000 images from the MS-COCO dataset and use a random object in the image to produce descriptions with different lengths. We report the average length of the predicted caption, and the fraction of times the predicted caption has a length equal to the target length in Table~\ref{tab:fidelity}. We find \methodname's outputs are mostly compliant with the target length.

\begin{table}[t]
    \begin{subtable}[c]{0.42\textwidth}
    \centering
    \footnotesize
    \begin{tabular}{l|c}
    Method & mAP ($\uparrow$)\\
    \midrule
CLIP~\cite{wu2024clim} & 29.2\\    
CLIM~\cite{wu2024clim}  & 62.2\\
RegionCLIP~\cite{zhong2022regionclip} & 62.8 \\       
\methodname (Top-1 caption) & 72.0\\
\methodname (Top-20 captions) & \textbf{85.0}\\
    \end{tabular}
     \caption{\textbf{MS-COCO Region Classification} with ground truth bounding box.}
    \label{tab:region_classification}
    \end{subtable}
    ~~~~
\begin{subtable}[l]{0.58\textwidth}
\footnotesize
    \centering
 \begin{tabular}{c|cc||c|cc}
         Target  & Mean Len & Accu- & Target  & Mean Len & Accu- \\
          Length &  of Pred. & racy & Length &  of Pred.  &racy\\
         \midrule
1 & 1.02 & 0.99 & 5 & 5.06 & 0.94\\
2 & 2.05 & 0.96 & 6 & 6.01 & 0.97\\
3 & 3.06 & 0.95 & 7 & 7.02 & 0.95\\
4 & 4.04 & 0.96 & 8 & 8.02 & 0.95\\
    \end{tabular}
    \caption{\textbf{Compliance metrics.} \methodname produces length-compliant captions for different lengths.}
    \label{tab:fidelity}
    \end{subtable}
    ~
    \caption{\textbf{\methodname's outputs are accurate and length compliant.}}
    \vspace{-0.25in}
\end{table}
\subsection{Visual Question Answering}
\label{sec:vqa}

Visual question answering (VQA) often requires visually grounded rich semantic understanding of the content at multiple levels of granularity depending on the question. These properties make VQA a great test-bed for our method which can generate dense spatially grounded information on visual content with desired semantic complexity. 

\noindent\textbf{FlexCap-LLM.} In Figure~\ref{fig:teaser} (right), we show how we use FlexCap with an LLM to solve visual questions. First, we convert an image to a sequence of localized descriptions that describe the image in terms of the objects and regions present in the image. To do so, we need region proposals. We use OWL-ViTv2~\cite{minderer2023scaling} to localize important objects and regions in an image. We keep the top 128 bounding boxes by their \textit{objectness} scores. We then use \methodname to describe each box in the image in the context of the entire image. In order to produce holistic descriptions, we use multiple prefixes for each region. These prefixes are a combination of length conditioning token and some initial text. We add the boxes and their descriptions to a \textit{text preamble} as context to the LLM (see Figure~\ref{fig:teaser})  that defines the setup where we are using an LLM to answer questions about an image. In all the experiments, we use PALM2-S model~\cite{anil2023palm} as the LLM of choice. We refer to this end-to-end system that takes an image and a question to output the answer as \textit{FlexCap-LLM}.

To adapt the base \methodname to have improved detection skills, output longer sentences, and identify OCR, we co-train \methodname for 25k more steps on detection (COCO, VOC, OpenImages, LVIS), captioning (COCO Captions, Visual Genome) and OCR datasets (WebLI). For image captioning datasets, we use the bounding box that covers the whole image. 
We find this co-training step useful for downstream tasks using the LLM. 

Hence we evaluate the effectiveness of FlexCap-LLM on several image VQA benchmarks such as OKVQA~\cite{Marino2019OKVQAAV}, VQAv2~\cite{Goyal2016MakingTV}, 
GQA~\cite{Hudson2019GQAAN}, and VizWiz~\cite{Gurari2018VizWizGC}, and video question answering benchmarks such as MSRVTT~\cite{xu2016msr} and MSVD~\cite{xu2017video}. Diverse characterictics of these datasets helps gaining better insight on \methodname's capabilities. We report the commonly used accuracy metric for each dataset.

\begin{table}[t]
\begin{subtable}[l]{0.5\textwidth}
\centering
\begin{tabular}{@{}lc@{}}
Method                        & \multicolumn{1}{l}{Acc.(\%) $\uparrow$} \\ \midrule

{\color[HTML]{9B9B9B} PalmE-562B~\cite{driess2023palm}} & {\color[HTML]{9B9B9B} 80.0} \\
\color[HTML]{9B9B9B} VLMO~\cite{bao2022vlmo}   & {\color[HTML]{9B9B9B} 82.8}     \\
\color[HTML]{9B9B9B} BEIT-3~\cite{wang2022image} & {\color[HTML]{9B9B9B} 84.2}     \\
{\color[HTML]{9B9B9B} PaLI-17B~\cite{chen2022pali}}   & {\color[HTML]{9B9B9B} 84.3}       \\ \midrule
FewVLM~\cite{jin2021good}                        & 47.7                             \\
Flamingo~\cite{Alayrac2022FlamingoAV} & 56.3\\
BLIPv2~\cite{Li2023-px}                        & 65.2                       \\
FlexCap-LLM & \textbf{65.6}\\
\end{tabular}
\caption{\textbf{VQAv2 results} on test-dev set }
\label{tab:vqav2}
\end{subtable}

\begin{subtable}[l]{0.5\textwidth}
\centering
\begin{tabular}{@{}lc@{}}
Method                        & \multicolumn{1}{l}{Acc.(\%) $\uparrow$} \\ \midrule
\color[HTML]{9B9B9B} LGCN~\cite{hu2019language}   & {\color[HTML]{9B9B9B} 55.8}     \\
\color[HTML]{9B9B9B} LXMERT~\cite{tan2019lxmert} & {\color[HTML]{9B9B9B} 60.0}     \\
\color[HTML]{9B9B9B} NSM~\cite{hudson2019learning}    & {\color[HTML]{9B9B9B} 63.0}     \\
\color[HTML]{9B9B9B} CFR~\cite{nguyen2022coarse}    & {\color[HTML]{9B9B9B} 72.1}     \\ \midrule
BLIPv2~\cite{Li2023-px}                        & 44.7                             \\
ViperGPT~\cite{Suris2023-ja}                        & 48.1                             \\
FlexCap-LLM & 48.8\\
InstructBLIP~\cite{dai2024instructblip} & \textbf{49.5}\\
\end{tabular}
\caption{\textbf{GQA results} on test-dev balanced set }
\label{tab:gqa}
\end{subtable}

\begin{subtable}[c]{0.5\textwidth}
\centering
\begin{tabular}{@{}lc@{}}
Method                                       & \multicolumn{1}{l}{Acc.(\%) $\uparrow$}               \\ \midrule
{\color[HTML]{9B9B9B} PalmE-12B~\cite{driess2023palm}}  & {\color[HTML]{9B9B9B} 55.5}          \\
{\color[HTML]{9B9B9B} PalmE-562B~\cite{driess2023palm}} & {\color[HTML]{9B9B9B} 66.1} \\
{\color[HTML]{9B9B9B} PaLI-3B~\cite{chen2022pali}}         & {\color[HTML]{9B9B9B} 52.4}          \\
{\color[HTML]{9B9B9B} PaLI-17B~\cite{chen2022pali}}        & {\color[HTML]{9B9B9B} 64.5}          \\ \midrule
BLIPv2~\cite{Li2023-px}                                       & 45.9                                 \\
Flamingo~\cite{Alayrac2022-fm}                                 & 50.6                                 \\
ViperGPT~\cite{Suris2023-ja}                                     & 51.9                        \\
FlexCap-LLM & \textbf{52.1}\\
\end{tabular}
\caption{\textbf{OKVQA results} on val set}
\label{tab:okvqa}
\end{subtable}
\begin{subtable}[c]{0.5\textwidth}
\centering
\begin{tabular}{@{}lc@{}}
Method                                  & \multicolumn{1}{l}{Acc.(\%) $\uparrow$}            \\ \midrule
{\color[HTML]{9B9B9B} Flamingo 32 Shot ~\cite{driess2023palm}} & {\color[HTML]{9B9B9B} 49.8}       \\
{\color[HTML]{9B9B9B} Flamingo FT~\cite{driess2023palm}}      & {\color[HTML]{9B9B9B} 65.7}       \\
{\color[HTML]{9B9B9B} PaLI-3B~\cite{chen2022pali}}    & {\color[HTML]{9B9B9B} 67.5}       \\
{\color[HTML]{9B9B9B} PaLI-17B~\cite{chen2022pali}}   & {\color[HTML]{9B9B9B} 74.4}       \\ \midrule
Flamingo~\cite{Alayrac2022-fm}                            & 31.6                              \\
Emu~\cite{sun2024emu} & 34.2\\
InstructBLIP~\cite{dai2024instructblip} & 34.5\\
FlexCap-LLM & \textbf{41.8}\\
\end{tabular}
\caption{\textbf{VizWiz results} on test-dev set}
\label{tab:vizwiz}
\end{subtable}

\caption{{\bf Zero-shot image question answering results}. FlexCap-LLM  is compared against recent baselines. Grayed out methods are trained on question answering datasets.}
\vspace{-.25in}
\label{tab:image_vqa}
\end{table}

\begin{table}[!htb]
    \begin{minipage}{.35\linewidth}
      \footnotesize
      \centering
\begin{tabular}{@{}l|cc@{}}
                   & \multicolumn{2}{c}{Acc.(\%) $\uparrow$} \\
                   \midrule
    Method         & \multicolumn{1}{l}{MSRVTT} & \multicolumn{1}{l}{MSVD} \\ \midrule
Emu~\cite{sun2024emu} & 8.3 & 18.8\\
Flamingo~\cite{Alayrac2022-fm} &  17.4                      & 35.6                       \\           
FlexCap-LLM~~~     & {\bf 25.0}                      & {\bf39.5} \\
\\
\end{tabular}
\caption{\textbf{Zero-shot video question answering results} reported on MSRVTT-QA and MSVD-QA on the test set. FlexCap-LLM is better than other zero-shot baselines for video VQA benchmarks.}
\label{tab:video_vqa}
    \end{minipage}%
    ~~~~~~~
    \begin{minipage}{.6\linewidth}
    \begin{subtable}{.4\linewidth}
      \centering
      \footnotesize
\begin{tabular}{l|c}
         Method & mAP $\uparrow$\\
         \midrule
         FCLN~\cite{johnson2016densecap} & 27.11 \\
         CAG-Net~\cite{yin2019context} & 36.29 \\
         \methodname  &  \textbf{46.9} \\
        \bottomrule
    \end{tabular}
    \subcaption{Captioning GT Boxes}
    \label{tab:densecaptioning_gt}
    \end{subtable}%
    \begin{subtable}{.6\linewidth}
    \footnotesize
      \centering
\begin{tabular}{l|c}
         Method & mAP$\uparrow$\\
         \midrule
         FCLN~\cite{johnson2016densecap}  & 5.39\\
         JIVC~\cite{yang2017dense} &9.31\\

         COCG~\cite{li2019learning}  &9.82\\
         CAG-Net~\cite{yin2019context}&10.51\\
         TDC+ROCSU
         ~\cite{shao2022region}&11.49\\
         GRiT~\cite{wu2022grit}  & 15.52\\
         \midrule
         \methodname + GRiT Boxes  & \textbf{16.2} \\
        \bottomrule
    \end{tabular}
     \subcaption{Dense Captioning}
    \label{tab:densecaptioning}
\end{subtable}
\caption{\textbf{Captioning boxes in Visual Genome dataset.} \methodname exceeds performance of other methods. All methods have been fine-tuned on Visual Genome captions.}
    \end{minipage} 
    \vspace{-.25in}
\end{table}

\noindent\textbf{Image Question Answering.} First we evaluate \methodname-LLM on VQAv2 , GQA, OKVQA and VizWiz image VQA benchmarks in a zero-shot setting, meaning that our approach is not trained with the task or the corresponding dataset. The results on these benchmarks are presented in Table~\ref{tab:image_vqa}.

\noindent\textit{Standard VQA}. The VQAv2 dataset is a standard for evaluating the performance of visual question-answering systems. In Table~\ref{tab:vqav2}, we present the results of our evaluation of FlexCap-LLM on this dataset. We find that FlexCap-LLM outperforms other zero-shot baselines, such as BLIP-2. This performance is achieved by providing object and region level information to LLMs without requiring multi-modal fine-tuning.

\noindent\textit{Compositional VQA}. The GQA dataset is for evaluating the performance on complex compositional questions. As \methodname produces information for multiple visual elements in the scene with their corresponding locations, it is quite well-suited for questions on compositional understanding of the image. On this benchmark, as shown in Table~\ref{tab:gqa}, FlexCap-LLM outperforms all the recent baselines except InstructBLIP~\cite{dai2024instructblip}.

\noindent\textit{VQA with External Knowledge}. OKVQA dataset is particularly designed for evaluating the ability to answer questions about images that require external knowledge which is not readily available on the image. Hence it requires multiple levels of understanding of the content, and reasoning with that information, which is  well-suited for applying \methodname.
In Table~\ref{tab:okvqa} we show our performance on OKVQA is superior to strong baselines such as Flamingo and ViperGPT which highlights the effectiveness of the mix of generic and specific descriptions generated by FlexCap. Unlike other baselines which use the question, \methodname generates captions without having access to the question.

\noindent\textit{VQA with atypical images}. We also evaluate on VizWiz, which contains visual questions asked by people who are visually impaired. Unlike web content, in these images the objects and the scene are not always well-centered, hence this dataset contains many out-of-distribution samples compared to typical web-crawled datasets. We report the results of this experiment in Table~\ref{tab:vizwiz}. Nevertheless, our approach significantly 
outperforms Flamingo~\cite{Alayrac2022FlamingoAV} and InstructBLIP~\cite{dai2024instructblip} in the zero-shot setting.

\noindent\textbf{Video Question Answering.} We also evaluate FlexCap-LLM on zero-shot video question answering datasets MSRVTT-QA and MSVD-QA~\cite{xu2017video}. The results on these benchmarks are presented in Table~\ref{tab:video_vqa}.
For processing the video, we sample 8 frames uniformly from the video. We pass each of these frames through FlexCap to produce captions of objects and regions. We then combine all the object captions from the different frames into one prompt for the LLM. 
We observe FlexCap-LLM significantly exceeds the performance of the Flamingo model in the zero-shot setting. These results highlight the zero-shot effectiveness of our method, which can solve tasks in the video domain even though both \methodname and the LLM were not trained for those tasks.

\subsection{Dense Captioning}
\label{sec:dc}
\textbf{Dataset and Evaluation Metrics.} The dense captioning task is defined as producing both the regions and the corresponding descriptions for each region. For this experiment, we use the Visual Genome~\cite{krishna2017visual} dataset. In this dataset, each image is annotated with multiple bounding boxes and each box has a corresponding caption. We use the train-test splits and evaluation metric as proposed in~\cite{johnson2016densecap}. The paper proposes to use a mean of Average Precisions (mAP) over pairwise thresholds of both IOU thresholds (0.3, 0.4, 0.5, 0.6, 0.7) and Meteor score thresholds (0.0, 0.05, 0.1, 0.15, 0.2, 0.25). We use the same preprocessing of text and boxes as mentioned in~\cite{wu2022grit}. 

\noindent\textbf{Fine-tuning \methodname.} We fine-tune the pretrained \methodname model on the Visual Genome train split for 60k steps with a lower learning rate of $1e-6$ at a resolution of $448 \times 448$.

\noindent\textbf{Captioning GT boxes}. 
Following the evaluation procedure from~\cite{johnson2016densecap}, we evaluate captioning of the ground-truth boxes in Visual Genome. Since this setting removes the localization task, we have a cleaner evaluation of only the region captioning problem. The results of this experiment are provided in Table~\ref{tab:densecaptioning_gt} in which we show that \methodname achieves better performance compared to other approaches evaluated in this setting.

\noindent\textbf{Captioning GRIT boxes}. 
In this experiment, we want to compare against other approaches that perform both localization and captioning. We are measuring how well \methodname performs when provided with box proposals.  Table~\ref{tab:densecaptioning}  shows \methodname obtains better performance compared to other approaches evaluated in this setting. Since in our work we do not propose any localization module, we use GRIT's~\cite{wu2022grit} region proposals as the input boxes for our model. This also allows us to directly compare our captioning capabilities against GRIT. We find that our approach outperforms GRIT at this task even though we test at a lower resolution of $448 \times 448$.

\subsection{Open-Ended Object Detection}
\label{sec:det_desc}
\begin{figure*}[t!]
\footnotesize{
    \centering
    \begin{subfigure}[t]{0.18\textwidth}
        \centering
        \includegraphics[height=.95in]{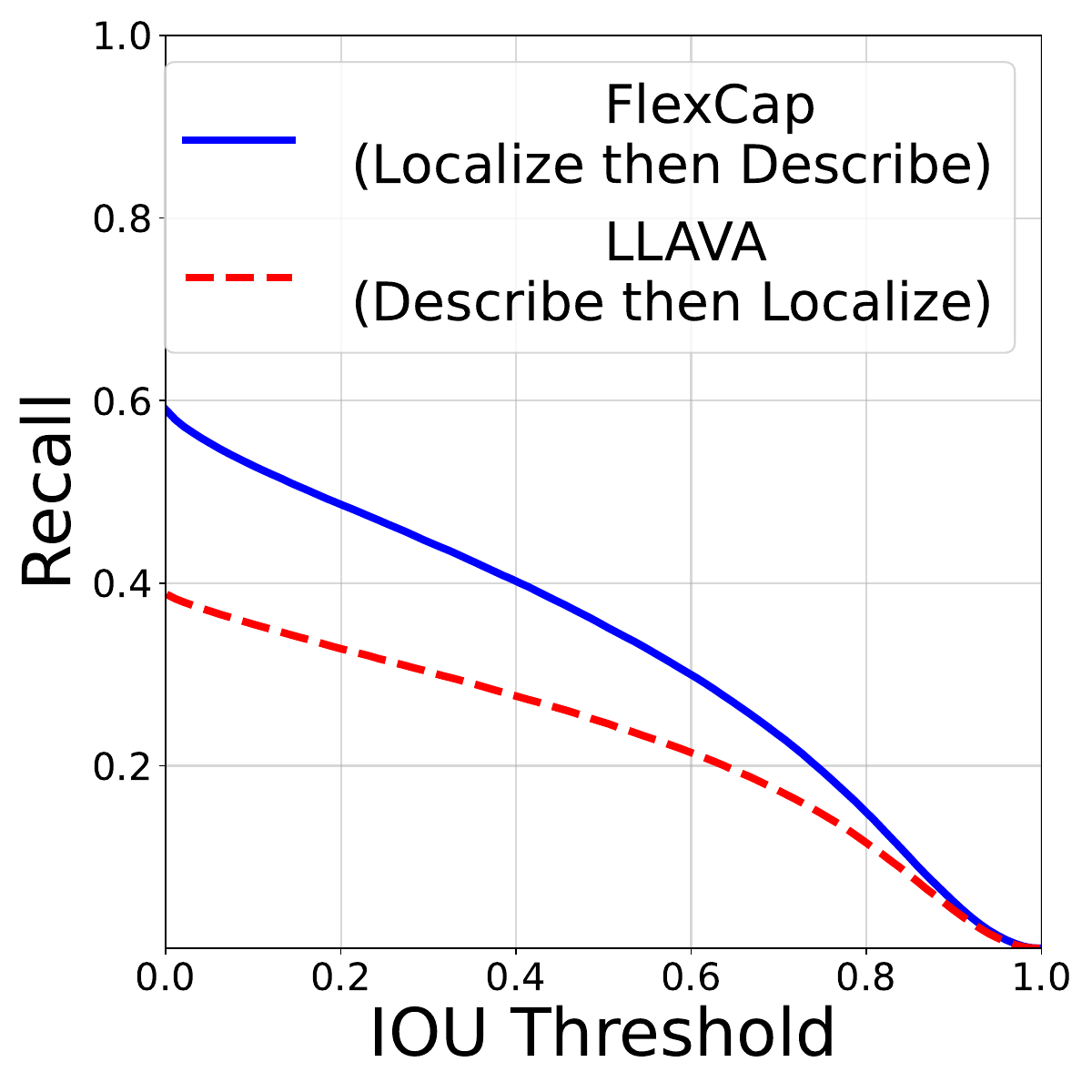}
        \caption{Recall v/s IOU}
    \end{subfigure}%
    ~ 
    \begin{subfigure}[t]{0.18\textwidth}
        \centering
        \includegraphics[height=.95in]{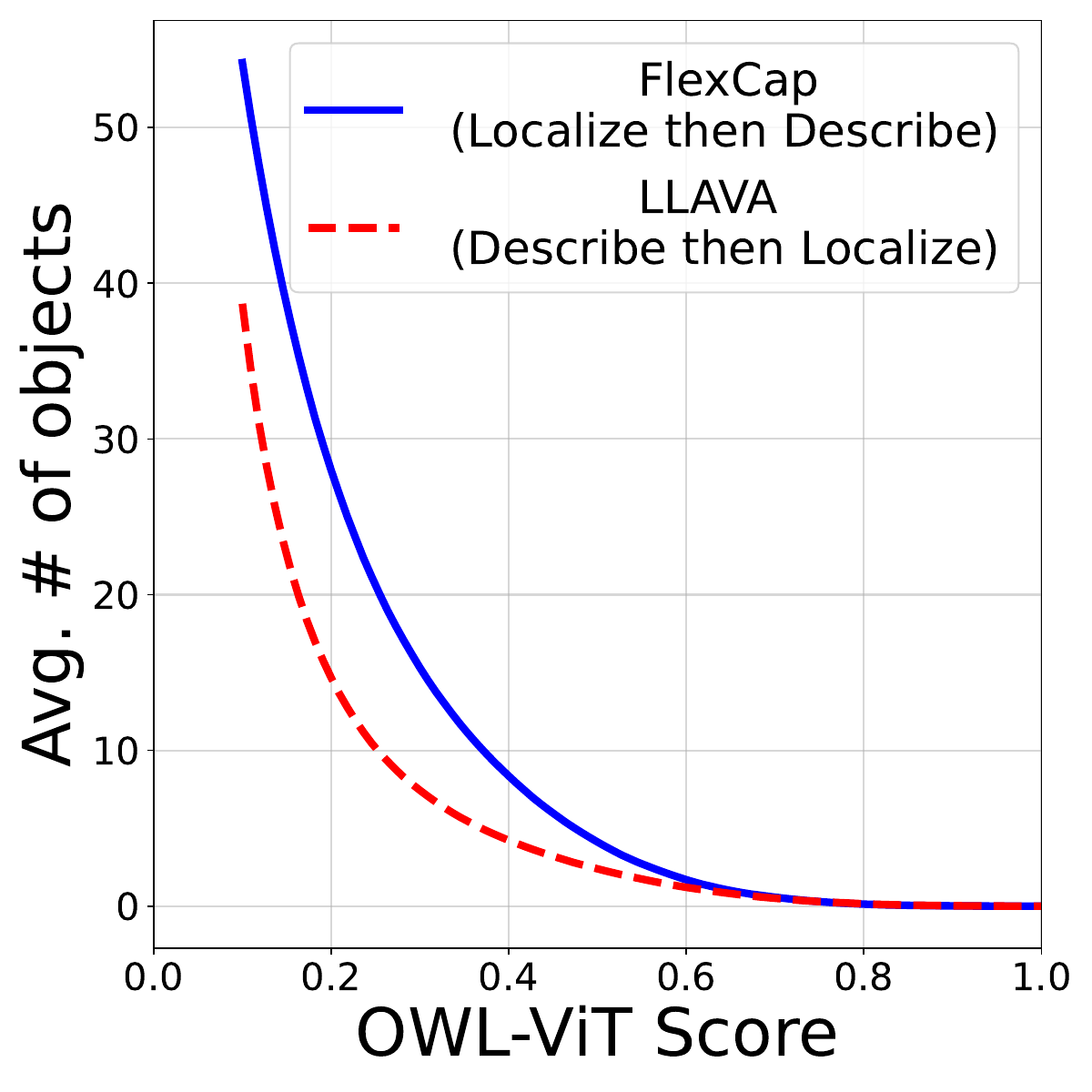}
        \caption{Number of\\Objects Found}
    \end{subfigure}
    ~
    \begin{subfigure}[t]{0.18\textwidth}
        \centering
        \includegraphics[height=.95in]{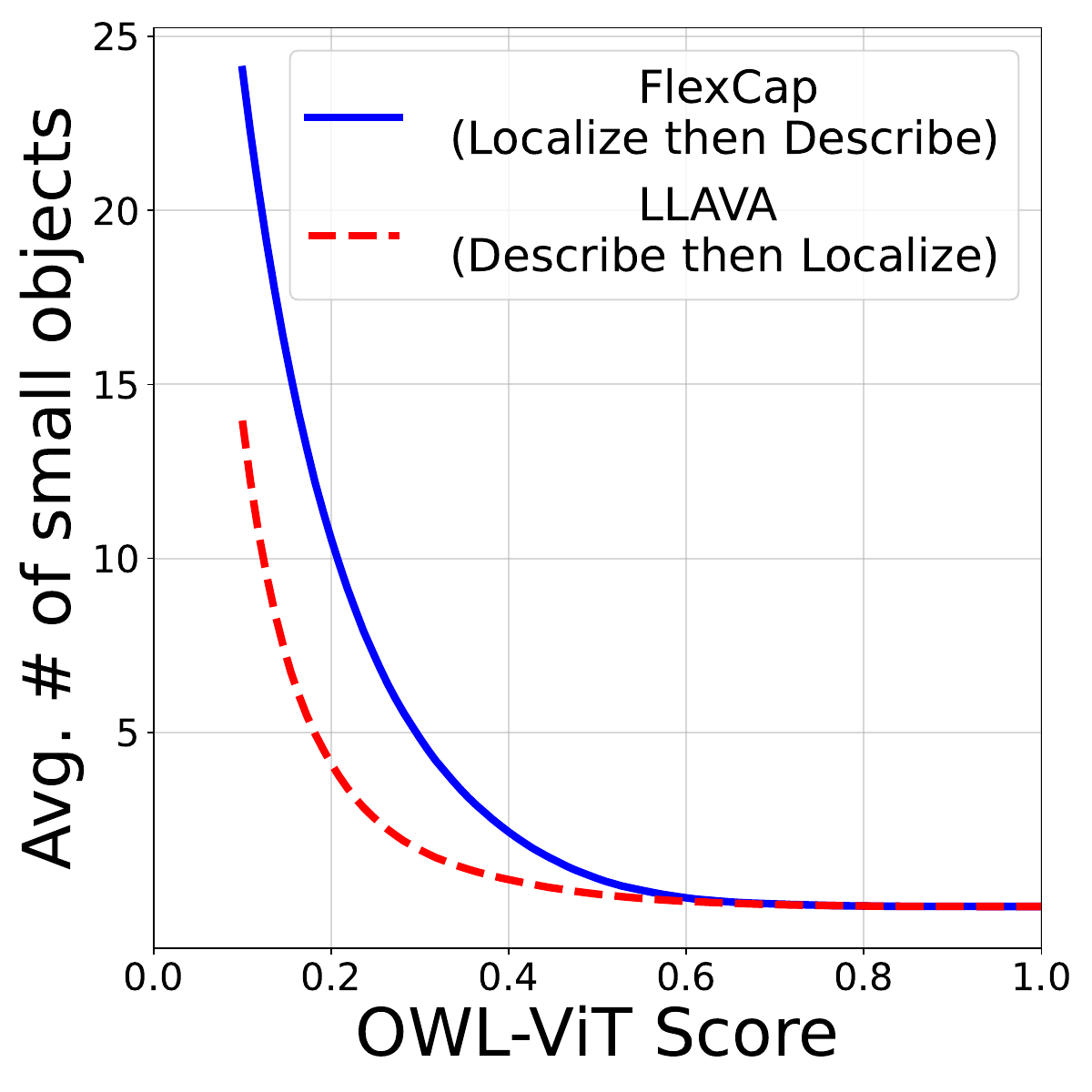}
        \caption{Number of Small Objects Found}
    \end{subfigure}
    ~
    \begin{subfigure}[t]{0.18\textwidth}
        \centering
        \includegraphics[height=.95in]{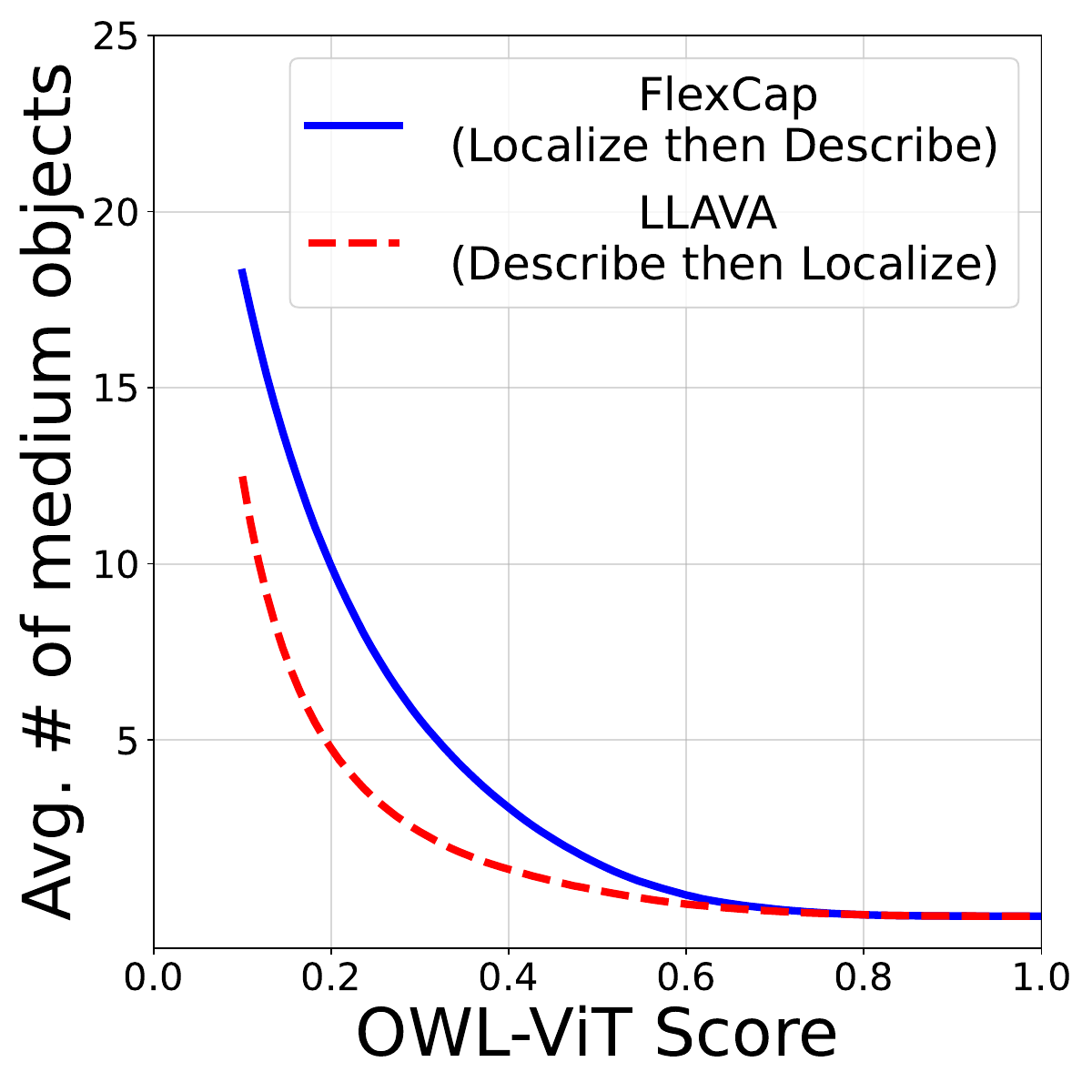}
        \caption{Number of Med. Objects Found}
    \end{subfigure}
    ~
    \begin{subfigure}[t]{0.18\textwidth}
        \centering
        \includegraphics[height=.95in]{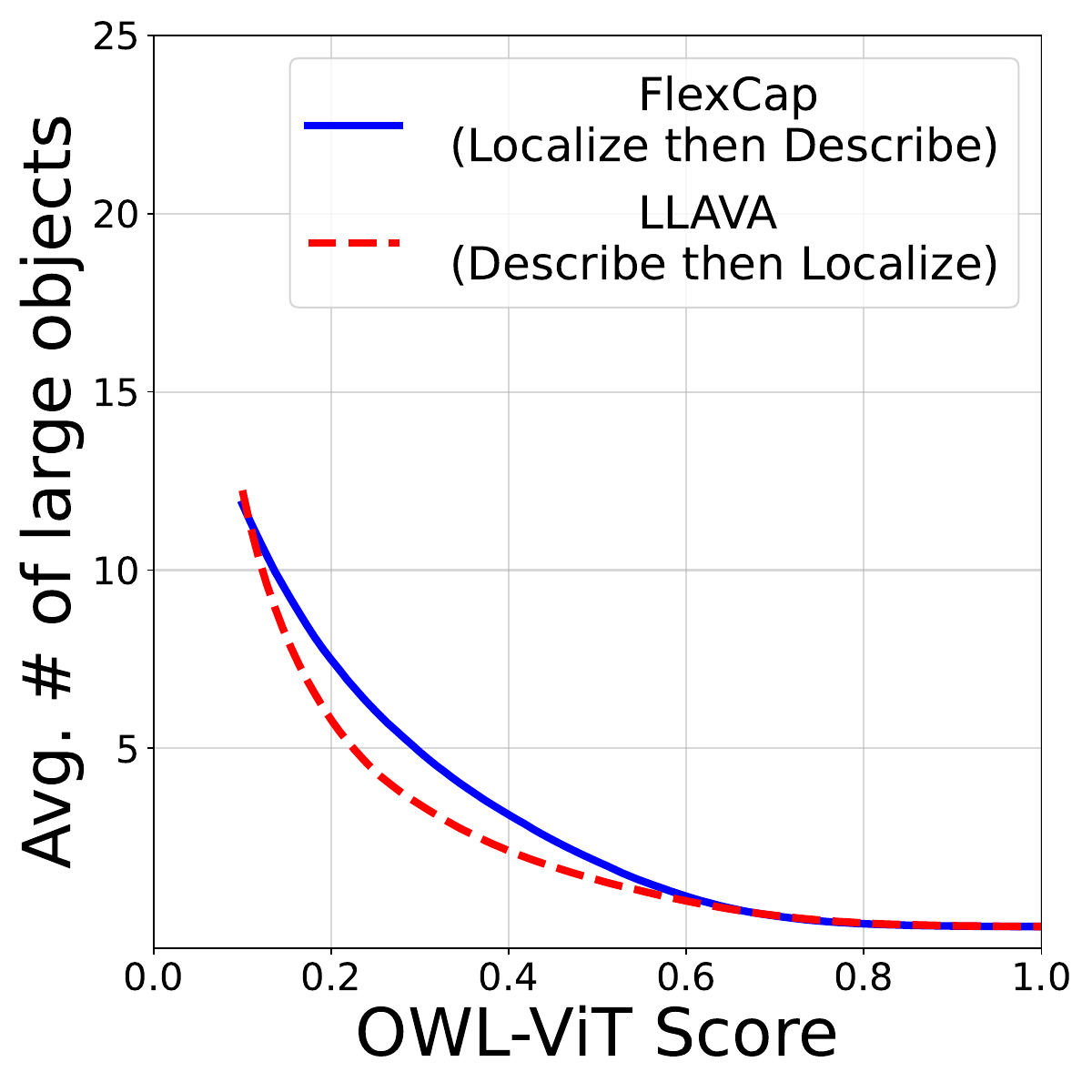}
        \caption{Number of Large Objects Found}
    \end{subfigure}
    \caption{\textbf{Open-Ended Object Detection} on the Visual Genome dataset. We find that the localize-then-describe approach, which involves describing every detected box with FlexCap, achieves higher recall (see (a)) and produces more bounding boxes with good matching scores (see (b)) compared to describe-then-localize approach with SOTA VLMs and open-vocabulary detection. The difference between the two approaches is most stark for small and medium sized objects (see (c-e)).
    }
    \label{fig:det_desc}
    }
    \vspace{-.18in}
\end{figure*}

Open-ended object detection, that is identifying all the objects in the image with rich descriptions, can be achieved in two ways: (a) localize-then-describe or (b) describe-then-localize. With FlexCap, we employ a \emph{localize-then-describe} approach where boxes are generated with a box proposal mechanism (i.e. OWL-ViT) and then described by FlexCap. We compare these results with a \emph{describe-then-localize} approach where we first ask a state-of-the-art (SOTA) VLM (i.e.\ LLAVA~\cite{liu2023llava}) for a comprehensive description of the image with all objects and then localize these descriptions using a SOTA open-vocabulary object detection method (i.e. OWL-ViT~\cite{minderer}). The \textit{Recall} of all the objects in the image is a major criteria in open-ended object detection. As shown in Figure~\ref{fig:det_desc}(a), we demonstrate that \textit{localize-then-describe} approach powered with FlexCap performs significantly better compared to \textit{describe-then-localize} approach using existing works.
In Figure~\ref{fig:det_desc} (b-e), we observe that our \textit{localize-then-describe} approach retrieves more parts of the image, particularly the small and medium size objects, compared to the \textit{describe-then-localize} approach. This is mainly because most VLMs are trained to describe the salient objects in the image rather than exhaustively list all the objects present in the scene.

\begin{figure}[h!]
\begin{center}
  \includegraphics[width=\linewidth]{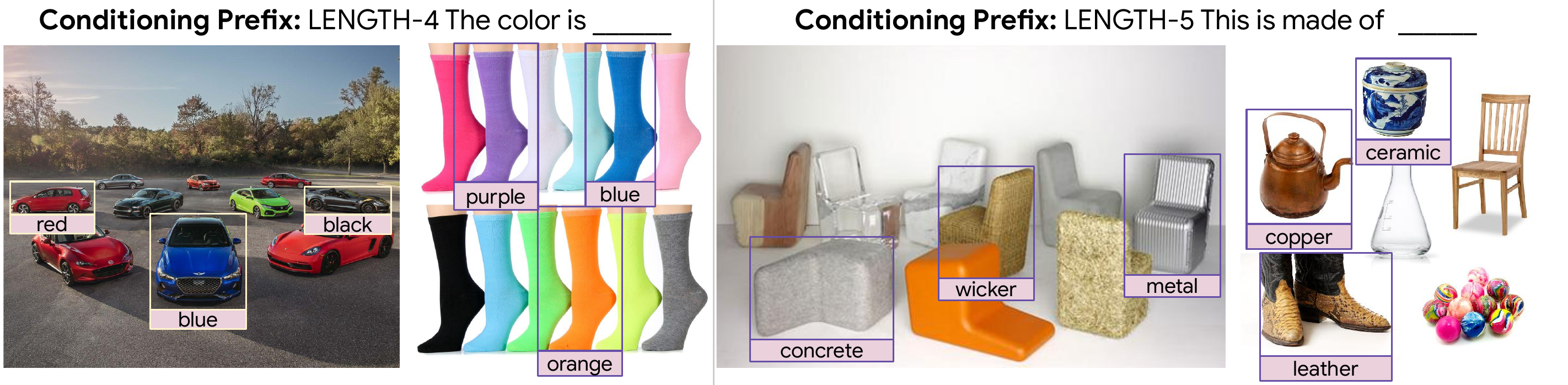}
\end{center}
  \caption{\textbf{Extracting properties by conditioning \methodname with prefixes.} Examples of \methodname extracting properties of objects of different categories by using relevant prefixes. Note how we are able to retrieve a one-word answer from the model by controlling the length of the caption.}
\label{fig:attributeprefix}
\vspace{-.15in}
\end{figure}

\begin{figure}[t!]
    \centering
    \includegraphics[width=\linewidth]{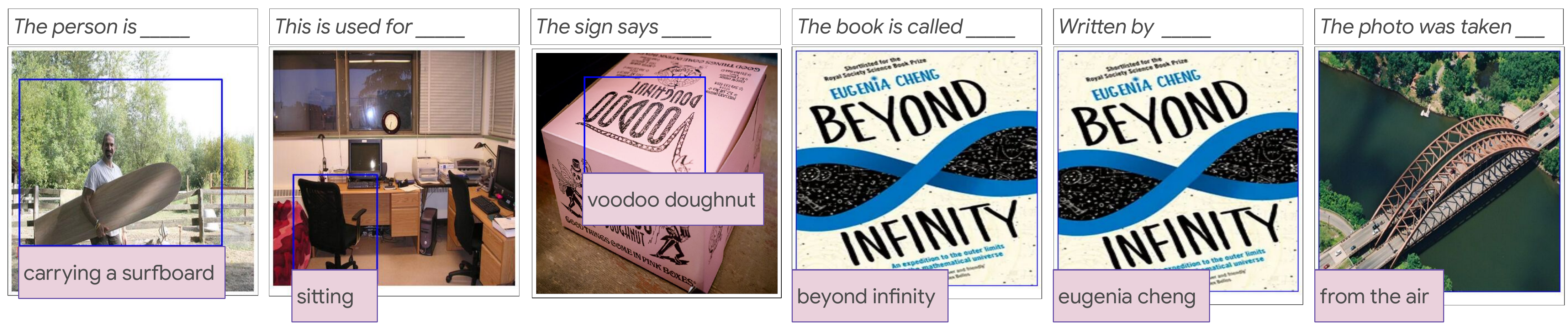}
    \caption{\textbf{Generating conditional captions using prefixes.} Conditional captions allow us to extract desired information from the input bounding box.}
    \label{fig:attribute_caption_collage}
    \vspace{-.15in}
\end{figure}

\subsection{Discussion}

\textbf{Prefixing.} Training on a large dataset also enables the extraction of different kinds of information from an image using \textit{prefixes}. 
We can use this property to our advantage to extract object properties such as color and material. We show examples of this in Figure~\ref{fig:attributeprefix}. Say we want to extract the color and material of objects, we design two prefixes: 1) "\textit{LENGTH-4 The color is }\verb|____|" and 2) "\textit{LENGTH-5 This is made of} \verb|____|". The pre-fixed length tokens guide the model to produce outputs that are just one more word. Note how the model is aware of  the input prefix and completes based on the object in the bounding box without confusing it with other surrounding objects. 

We show more examples of using the prefix to define  in Fig.~\ref{fig:attribute_caption_collage}. \methodname can extract the following attributes with the corresponding prefix in the brackets: the actions of humans (\textit{"The person is"}), the uses of objects (\textit{"This is used for"}), OCR (\textit{"The sign says"}), names of books (\textit{"The book is called"}), authors of books (\textit{"Written by"}), and  the locations where the photos were taken (\textit{"The photo was taken in"}). Note the same model can extract different information from the same bounding box: the name of the book title and the author name based on the prefixing (col 4-5).

\textbf{Limitations.} We use OWL-ViT and alt-text to generate the training dataset for the model rather than relying on human annotations. While this approach allows us to scale, there exist biases in the model and alt-text which will reflect in the outputs of FlexCap. Another limitation is that the proposed  FlexCapLLM system is not end-to-end trainable but is a composition of a VLM and LLM. This can be alleviated by training a VLM with the localized captioning dataset.

\textbf{Broader Impact.} Localized and controllable image captioning enabled by our model may benefit applications like accessibility tools for visual impairments and intuitive human-computer interaction. However, biases in training data raise concerns around potential misrepresentation or exclusion of certain demographics. Dual-use risks also exist if this model is employed for unethical surveillance. 

\section{Related Work}

\noindent {\bf Visual question answering} (VQA), a task designed to assess if a computer can answer questions about an image, often requires grounding visual concepts and reasoning. Although initially introduced for supervised evaluation of the task~\cite{antol2015vqa}, most recently, VQA also become one of the most powerful benchmarks for evaluating task and dataset independent visual dialog. Several existing models such as ViperGPT~\cite{Suris2023-ja}, Flamingo~\cite{Alayrac2022FlamingoAV}, BLIP~\cite{li2022blip,Li2023-px}, PaLI~\cite{chen2023pali} show convincing zero-shot performance on the VQA benchmarks that rivals the supervised approaches. Unlike most previous zero-shot approaches, which tightly couple vision and language components in a single model, \methodname generates a high-level human interpretable representation of an image and demonstrates that, through straight-forward application of LLMs, we can achieve comparable performance with state-of-the-art results across VQA benchmarks. Unlike others, ViperGPT~\cite{Suris2023-ja}, also decouples vision and language components and reinterprets visual questions with LLM generated programs, and executes them using existing visual perception tools. Whereas, in our case we use only one powerful vision tool, i.e. \methodname, to generate all the necessary information and leave the reasoning to an LLM. In that sense, \methodname is quite complementary to ViperGPT as it can be one of the powerful tools that can improve the controllable visual understanding of the image for ViperGPT.

\noindent \textbf{Open vocabulary object detection} models like \ OWL-ViT~\cite{minderer} and ViLD~\cite{gu2021open} enable the user to query any given text on the image and obtain matched bounding boxes for those queries. In these models the text is often encoded by a text encoder like CLIP~\cite{radford2021learning} and T5~\cite{raffel2020exploring}. The text embeddings are compared with the category-agnostic box proposals coming from the visual backbone. In this work, we use OWL-ViT's text and vision encoders to associate bounding boxes with text-queries to produce our training data. By training a localized captioning model, we remove the manual step of providing per-dataset or per-image text queries to use OWL-ViT. 
RegionCLIP~\cite{zhong2022regionclip} obtained good performance on open-vocabulary object detection by utilizing region-level vision-language contrastive learning on large scale data. We differ from this work as we generate the description for each bounding box instead of associating  text queries (defined manually) with bounding boxes.

\noindent {\bf Dense captioning} involves localizing salient regions of the image and describing them with natural language sentences, introduced in \cite{johnson2016densecap}. 
In practice, the existing work often produces longer and more informative descriptions of objects or their compositions using visual attributes of objects \cite{yin2019context,kim2019dense} or contextual and global image cues \cite{yang2017dense,li2019learning}. However, the richness of descriptions in this line of work are often limited to existing image captioning datasets \cite{lin2014microsoft,krishna2017visual}. By utilizing a large scale dataset  of billions of noisy image-text pairs collected from the web (similar to \cite{jia2021scaling,chen2022pali}), we aim to generate more diverse sentences with a focus on describing the visual content in controllable detail using a richer visual descriptive space learned from the web.

\noindent \textbf{Length-controlled image captioning} has been explored in ZeroCap~\cite{tewel2022zerocap} and LIC~\cite{deng2020length}. ZeroCap~\cite{tewel2022zerocap} implements length control as a post-processing step by changing the probability of sampling the end-of-sentence token. Hence the model is not naturally trained with word length conditioning in mind and cannot guarantee fine-grained length control at the level of number of words. On the other hand, LIC~\cite{deng2020length} generates length-controllable captions by conditioning the model with learned tokens that represent different length intervals. However there are considerable differences compared to \methodname. First, our approach allows for controllability at the level of image regions, while LIC only provides full image captions. This is a significant difference, as it allows us to generate concise or detailed captions for all the objects in
the image. Second, our approach has a more precise level of caption-length control. LIC uses a coarse subjective level
of control with four or five levels of length (e.g.\ short, medium, long, and longer), while our approach allows for an exact number of words to be specified. \cite{wang2023caption} also propose an approach to produce variable length descriptions for objects localized interactively. They use a pre-trained image captioner to produce descriptions of objects and ChatGPT in post-hoc to output length-conditioned captions. While in our \methodname model, the length tokens modulate the output produced by the captioner.
\section{Conclusion}

In this work, we introduce FlexCap, a flexible captioning model that can describe localized regions in an image with controllably rich captions. To train \methodname, we generate a large-scale image-box-caption dataset that is rich in diversity of visual descriptions and their length. We achieve this by utilizing existing web-scale noisy image-text pairs and open-vocabulary object detection models. 
We show how localized rich descriptions provided by FlexCap can help us connect images and videos to LLMs and achieve strong performance on visual question answering and dense captioning tasks. 
We also show that our \methodname model benefits from contrastive pretraining, localized captions dataset size scaling,  model size scaling, and is compliant to length conditioning. 
We also demonstrate the effectiveness of \methodname-enabled \textit{localize-then-describe} approach over the \textit{describe-then-localize} approaches.

\section*{Acknowledgements}

The authors would like to thank Matthias Minderer, Lisa Anne Hendricks, Andy Zeng, Matthias Bauer, Anastasija Ilić, Alexey Gritsenko, Vincent Vanhoucke, Jonathan Hoech, Alex Irpan, and Shefali Umrania for their feedback and helpful discussions.
{
    \small
    \bibliographystyle{abbrv}
    \bibliography{main}
}

\newpage
\appendix
\section*{Appendix}

\section{Localized Captions Dataset Details} 
\begin{figure}[h!]
\centering
 \begin{subfigure}[t]{0.475\textwidth}
        \centering
        \includegraphics[width=0.9\linewidth]{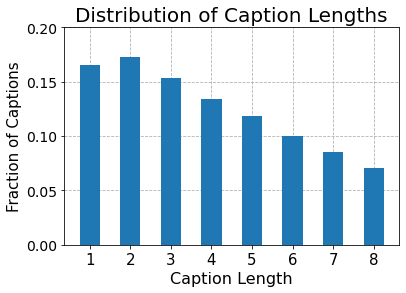}
        \caption{WebLI Dataset}
        \label{fig:caption_length_a}
    \end{subfigure}
  \begin{subfigure}[t]{0.475\textwidth}
        \centering
        \includegraphics[width=0.9\linewidth]{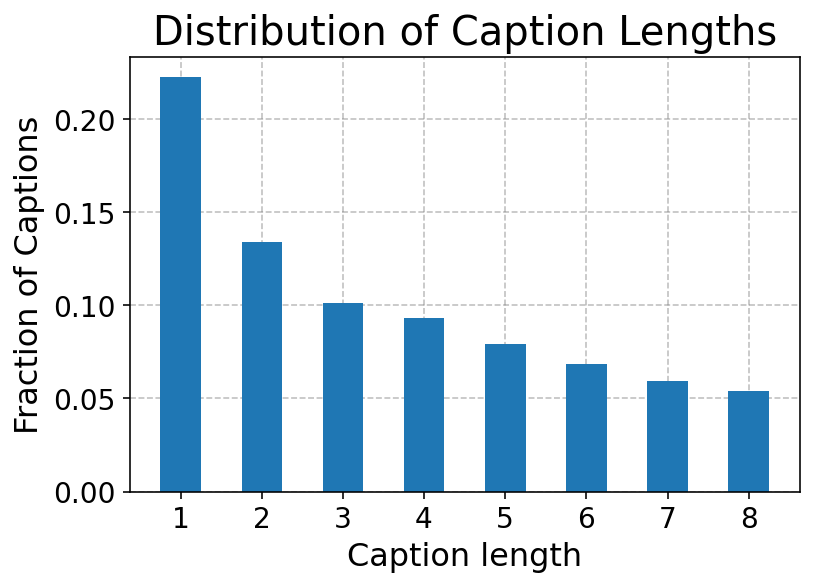}
        \caption{YFCC100M Dataset}
        \label{fig:caption_length_b}
    \end{subfigure}
   
   \caption{\textbf{Distribution of caption lengths in the Localized Captions Dataset.} We show the distribution of caption length for the region-captions obtained using paired image-texts from two datasets: WebLI~\cite{chen2022pali} and YFCC100M~\cite{thommeeyfcc100m}}.

\end{figure}

\noindent\textbf{WebLI.} Our dataset creation process on the WebLI dataset results in $\sim$32 billion Image-Box-Caption triplets. In Figure~\ref{fig:caption_length_a}, we show the distribution of caption lengths in the generated dataset using WebLI. We observe that the distribution is not uniform. This is due to the fact that there are more n-grams of length $1$ to sample than length $8$. The average number of unique boxes in an image is $4.19$, and the average number of captions per box is $4.04$.

\noindent\textbf{YFCC100M.}Our dataset creation process on the CLIP subset of the YFCC100M dataset results in $\sim$0.2 billion Image-Box-Caption triplets. In Figure~\ref{fig:caption_length_b}, we show the distribution of caption lengths in the generated dataset using YFCC100M. The average number of unique boxes in an image is $6.3$, and the average number of captions per box is $3.3$.

\begin{figure}[!htbp]
\begin{center}
   \includegraphics[height=7in]{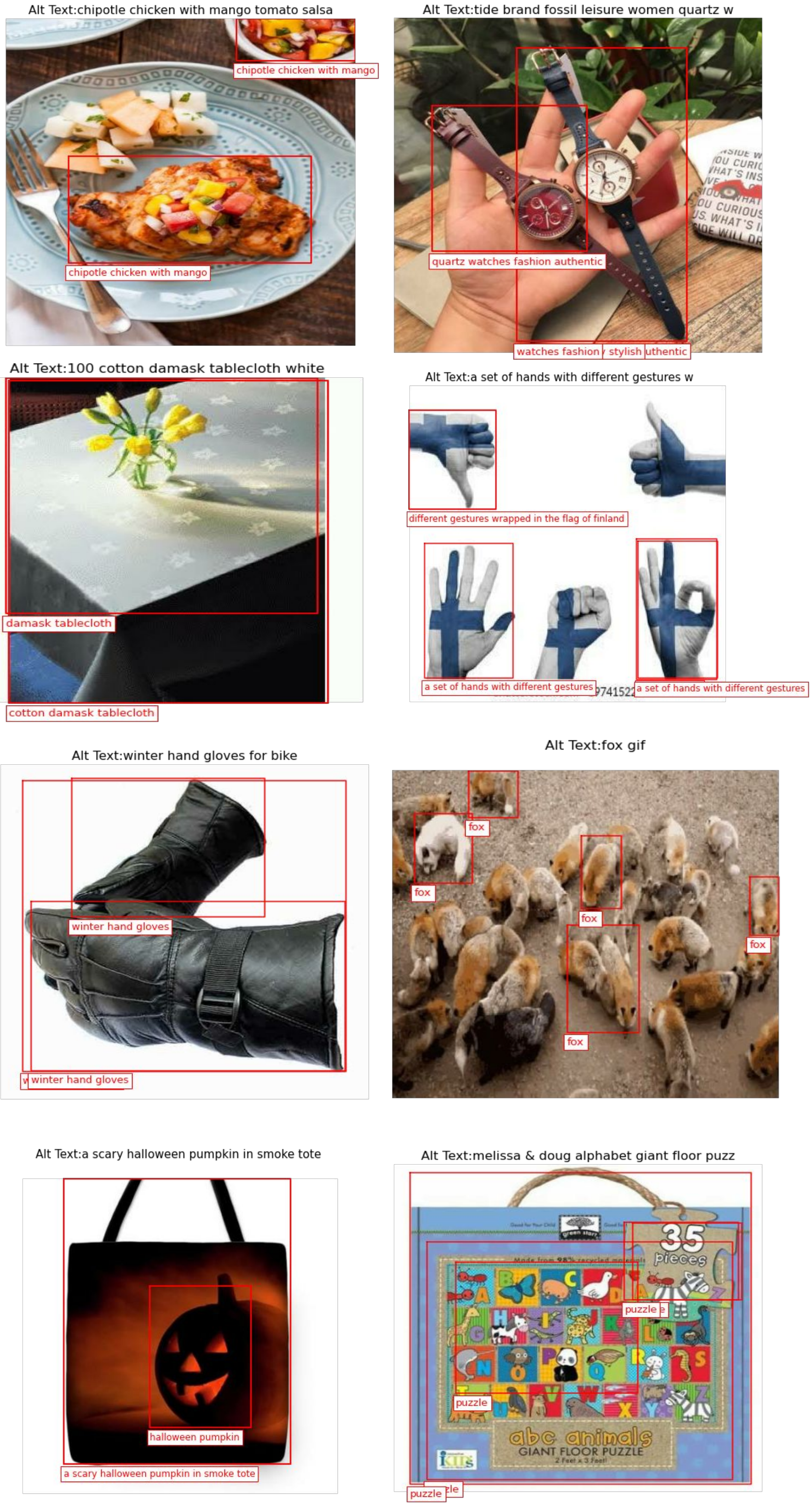}
\end{center}
   \caption{\textbf{Samples from the Localized Captions dataset with images from the WebLI dataset~\cite{chen2022pali}}. We only visualize a maximum of 5 boxes for each image to avoid clutter.
    }
\label{fig:dataset_samples}
\end{figure}

\noindent \textbf{N-gram Filtering.} Before matching n-grams with boxes, we filter out n-grams that do not form informative or grammatically correct captions for boxes. This is done with three steps: 1) Removing any captions composed only of uninformative words (image, jpg, background, wallpaper, hd wallpaper etc.); 2) Removing n-grams that begin with words with which sentences usually do not start (of, on, in etc.); 3) Removing n-grams that finish with words with which sentences usually do not end (a, the, to, on etc.). This step is essential to reduce noise present in the large-scale image-text pair dataset obtained from the web. It is also important for the captioning model to produce grammatically correct informative sentences.

\noindent\textbf{Dataset Samples.} We show some samples from the dataset in Figure~\ref{fig:dataset_samples}. The alt-text from which the box captions are generated is provided as the title of the image. Note the alt-text gets clipped due to display-length limits which is why the detected boxes might have captions not visible in the displayed alt-text directly. We next discuss how captions of varying lengths are matched with different objects in an image.

\section{Ablations and Analysis}

\label{sec:ablations}
\noindent\textbf{Large-scale Contrastive Pre-Training.} In this section, we study the impact of using a contrastively pre-trained vision encoder or training from scratch using only box-caption objective. For this experiment, we use ViT-B/16 as our backbone and train on the WebLI Region Captioning dataset. We report the results on ground truth (GT) box captioning  and dense captioning on GRiT proposals on the Visual Genome dataset in Table~\ref{subtab:cpt}. In Row 1, we observe that a length-conditioned region captioning objective can be used for training both the backbone and text decoder from scratch and achieve competitive performance. Second, we observe using a large-scale contrastively pre-trained (CPT) model results in better performance. Note that these CPT weights have already been open-sourced~\cite{zhai2023sigmoid}. 

\noindent\textbf{Data Scaling.} In this experiment we measure how scaling the localized captions dataset affects performance. We first take ViT-B/16 encoder contrastively pre-trained with a large image-text dataset (WebLI). We now train it on two different region captioning datasets of different scales. We show the results of this experiment in Table~\ref{subtab:data_scaling}. We find that even after contrastive pre-training on a large scale, a large region-caption dataset can result in significant performance boost ($\sim$5\% mAP for GT box captioning). This shows the importance of designing a scalable dataset creation method such as the one introduced in Section~\ref{sec:dc}. We also note that the dataset and pre-trained weights required to reproduce the model with YFCC100M are available publicly.

\noindent\textbf{Model Scaling.} Next, we measure the impact of changing the model size. We train two models: ViT-B/16 (85M) v/s SO-ViT/14 (428M). Just like the data scaling experiments, both of these have been pre-trained on the WebLI dataset in a contrastive manner. We report the results of this experiment in Table~\ref{subtab:model_scaling}. We find that the larger model results in better performance.

\begin{table}[t]
\small
\begin{subtable}[l]{0.20\textwidth}
\centering
\begin{tabular}{@{}c|cc@{}}
   & \multicolumn{2}{c}{VG mAP $\uparrow$} \\ 
   \midrule
    CPT         & \multicolumn{1}{c}{GT} & \multicolumn{1}{c}{GRiT} \\ \midrule
 & 43.6 & 15.6 \\
\checkmark & \textbf{45.1} & \textbf{15.8}\\
\end{tabular}
\caption{\textbf{Contrastive Pre-training}}
\label{subtab:cpt}
\end{subtable}
\begin{subtable}[l]{0.45\textwidth}
\centering
\begin{tabular}{@{}c|c|cc@{}}
 \multicolumn{2}{c|}{} & \multicolumn{2}{c}{VG mAP $\uparrow$} \\ 
\midrule
    Dataset   & Size &
   \multicolumn{1}{c}{GT} & \multicolumn{1}{c}{GRiT} \\ \midrule
YFCC100M & 0.2B & 38.5 & 14.2 \\
WebLI & 32B & \textbf{45.1 }& \textbf{15.8}\\
\end{tabular}
\caption{\textbf{Data Scaling}}
\label{subtab:data_scaling}
\end{subtable}

\begin{subtable}[r]{0.3\textwidth}
\centering
\begin{tabular}{@{}c|c|cc@{}}
 \multicolumn{2}{c|}{} & \multicolumn{2}{c}{VG mAP $\uparrow$} \\ 
\midrule
    Backbone    & Params     & \multicolumn{1}{c}{GT} & \multicolumn{1}{c}{GRiT} \\ \midrule
 ViT-B/16 & 86M &45.1 & 15.8 \\
SO-ViT/14 & 428M & \textbf{46.9}	& \textbf{16.2}\\
\end{tabular}
\caption{\textbf{Model Scaling}}
\label{subtab:model_scaling}
\end{subtable}
\caption{{\bf Ablations.} We vary different aspects of model pre-training, dataset size, and model size and measure its effect on region captioning task on the Visual Genome dataset.  In (a) and (b) the visual backbone is ViT-B/16 and in (b) and (c) all models start from contrastively pre-trained weights.}
\label{tab:ablations}
\vspace{-.25in}
\end{table}

\section{Open-Ended Object Detection}
\begin{figure*}[t!]
    \centering
        \includegraphics[width=\textwidth]{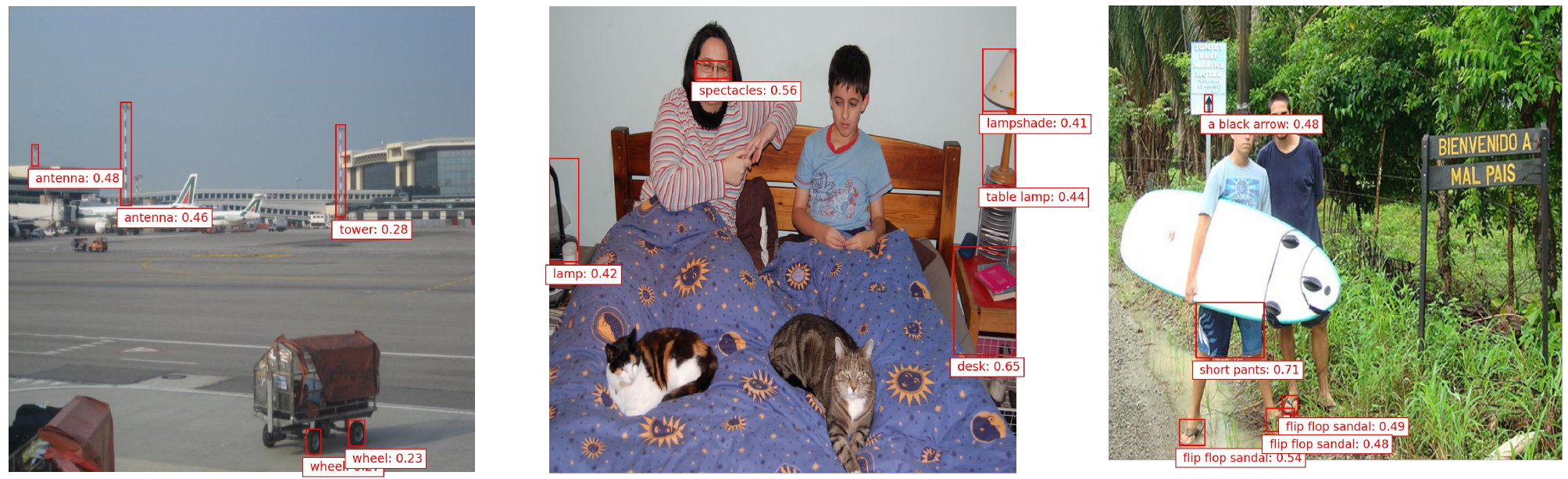}
    \caption{\textbf{Objects missed by LLAVA but found by FlexCap.} We show some  typical examples of objects (5 per image to avoid clutter) that FlexCap can find with OWL-ViT but LLAVA does not.}
    \label{fig:missed_det}
\end{figure*}

In this experiment, we compare the effectiveness of a \textit{describe-then-localize} method (specifically LLAVA 1.5~\cite{liu2023llava} 7B + OWL-ViTv2~\cite{minderer2023scaling}) with a \textit{localize-then-describe} method (specially FlexCap + OWL-ViTv2) for the task of open-ended object detection. For the \textit{describe-then-localize} method, we generate a list of objects using the following prompt:\verb|Describe object names and regions in this image|. We extract the nouns and use them as text queries for OWL-ViTv2. For the \textit{localize-then-describe} approach, we take the top 128 objects ranked by objectness score by OWL-ViTv2 and describe them with different length prompts. In Fig. 5 in the main paper, we compare recall of the ground-truth regions annotated in the Visual Genome dataset for both these approaches. We find that while \textit{describe-then-localize} with LLAVA can be an effective approach for finding large objects,  \textit{localize-then-describe} with FlexCap is  significantly better at medium and small sized objects and marginally better for large-sized objects.
We show some typical examples of missed detections in Fig.~\ref{fig:missed_det}.

\begin{figure}[h!]
    \centering
    \includegraphics[width=\linewidth]{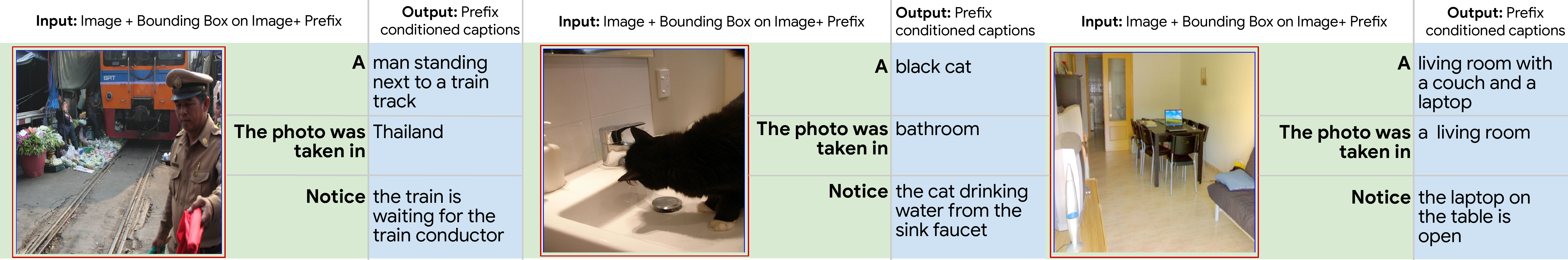}
    \caption{\textbf{Diverse captioning with Prefixes.} FlexCap can be used to perform conditional captioning of images. We show three prefixes: \textit{"a", "the photo was taken in ", "notice"} resulting in diverse captions for the same image. Note the input red bounding box is around the full image.}
    \label{fig:diff_prefixes}
\end{figure}

\section{Different prefixes with same image} 

In Fig.~\ref{fig:diff_prefixes}, we highlight how different prefixes can be used to generate diverse captions. Note how in column 1 the model correctly identifies the country even, possibly using the logo on the train. We find the prefix \textit{"Notice"} leads to captions highlighting noteworthy aspects in an image something which the un-prefixed caption does not do.

\section{Visual Dialog}
\begin{figure*}[t]
\begin{center}
   \includegraphics[width=1.0\linewidth]{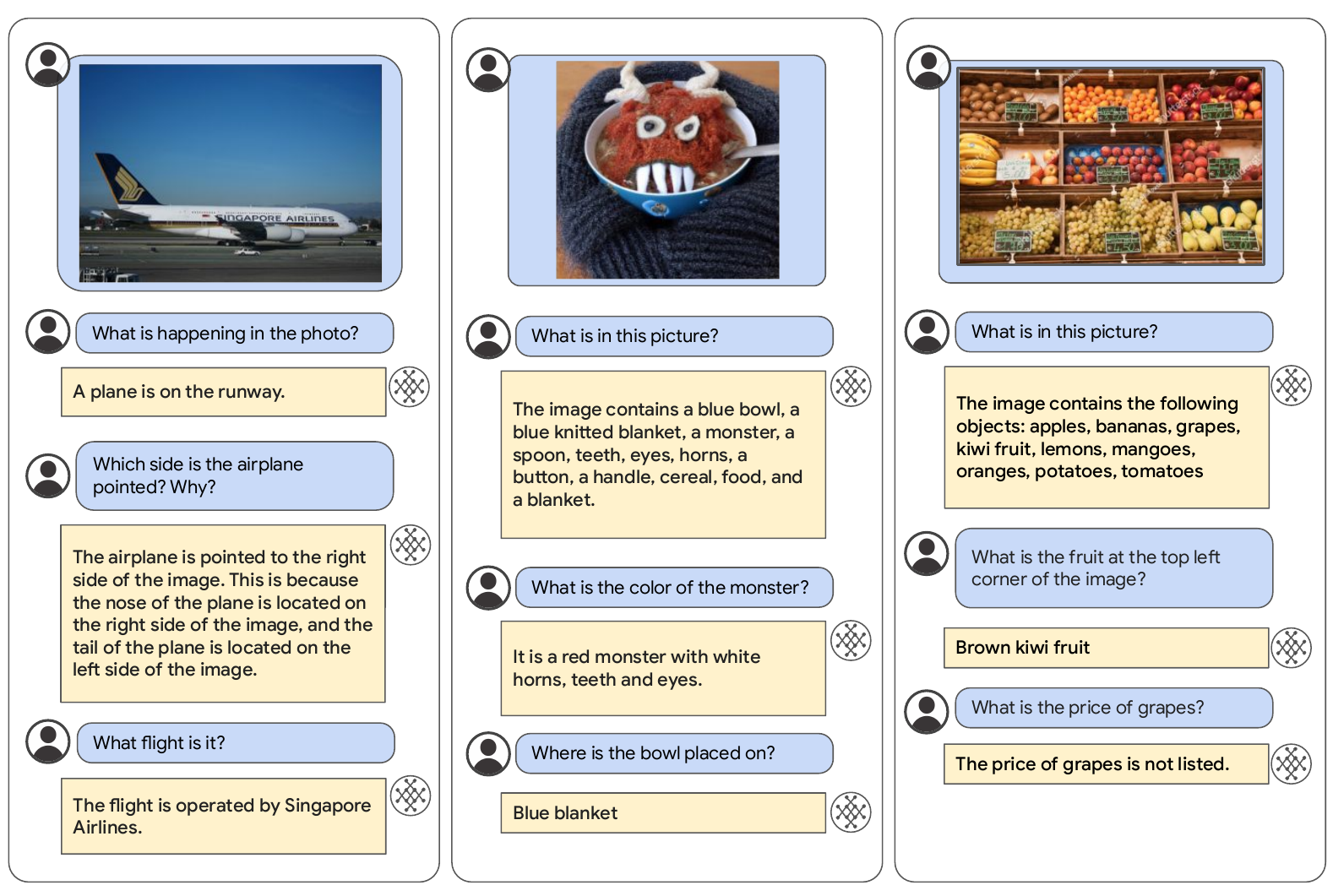}
\end{center}
   \caption{{\bf \methodname-LLM for Visual Dialog} }
\label{fig:vd}
\end{figure*}

\methodname-LLM can be used for the task of visual dialog~\cite{das2017visual}. We first caption all the objects in the image using \methodname and OWL-ViTv2. We retain the top 128 boxes according to the objectness score from OWL-ViTv2 and describe each box using FlexCap. See Fig. 1 and Fig. 4 in the main paper for more details. Once the image has been represented as the list of objects in text, we can interact with a LLM by providing the conversation turns as additional context for each query to the LLM. We show some examples of conversations with the \methodname-LLM system in Figure~\ref{fig:vd}. Note how the model is able to read text in the image in the leftmost figure, recognize material in the middle figure, and localize objects of interest in the rightmost figure. As we compute the object captions only once at the beginning of the conversation, there is no additional overhead of querying a large VLM for each additional turn in the conversation. We show more examples of this on the webpage in the FlexCap-LLM section of \verb|index.html|.

\begin{figure}[t]
    \centering
    \includegraphics[width=\textwidth]{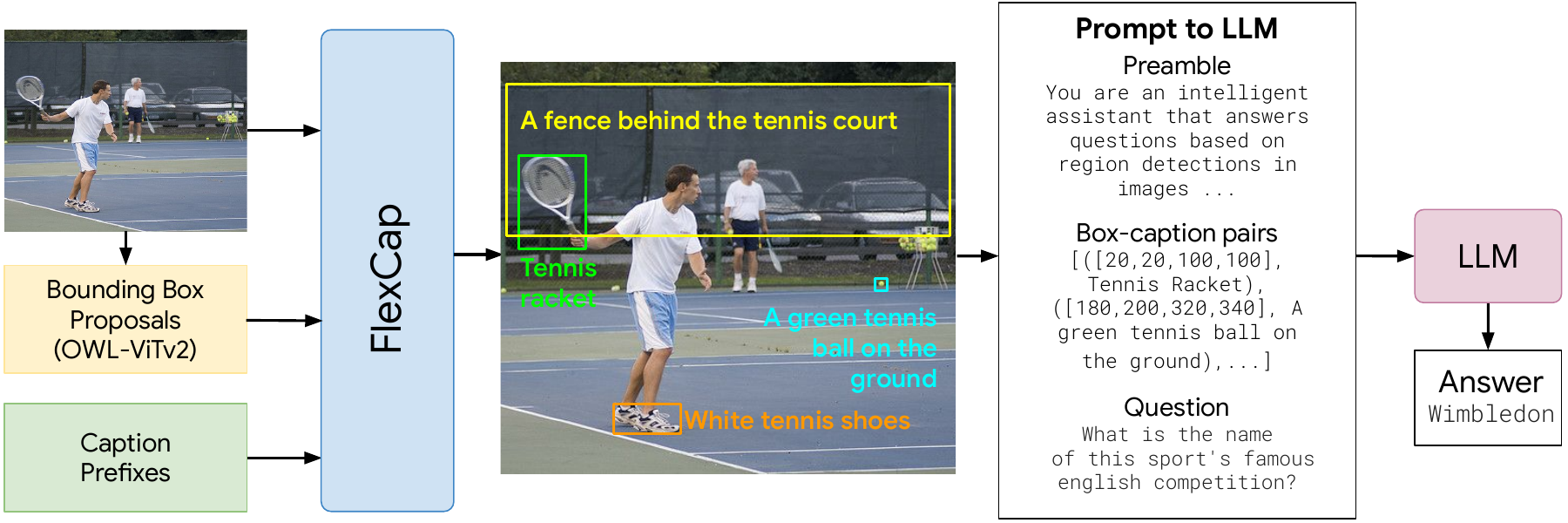}
    \caption{\textbf{\methodname for VQA with bounding box proposals and an LLM}. \methodname generates captions for different regions in a given image. To answer any open-ended questions, we prompt an  LLM~\cite{palm2} with \methodname's detections (box-caption pairs).}
    \label{fig:flexcap_with_llm}
\end{figure}

\section{Effect of Length on FlexCapLLM}
In this experiment, we study the effect of generating captions of different lengths on downstream visual question answering tasks. We evaluate this by using captions of increasing lengths on the val split of the VQAv2 dataset. We report the results of this experiment in Table~\ref{tab:length_ablation}. We find that increasing the length of generated captions results in better VQAv2 and GQA performance.

\begin{table}[h!]
\centering
\begin{tabular}{@{}l|cccc@{}}
   & \multicolumn{4}{c}{\textbf{Maximum Length of Captions}}\\
     \midrule
    \textbf{Dataset} & 1 & 2 & 4 & 8 \\ \midrule
     VQAv2 (\textit{val} split) & 57.4 & 59.9 & 66.4 & \textbf{67.8} \\
     GQA (\textit{testdev-balanced} split) & 43.6 & 45.3 & 47.7 & \textbf{48.8} \\
\end{tabular}
\caption{\textbf{Effect of caption lengths on FlexCap-LLM}}
\label{tab:length_ablation}
\end{table}

\section{FlexCapLLM Details} 
In Fig.~\ref{fig:flexcap_with_llm} we show the whole pipeline how \methodname interfaces with LLMs. We convert an image into a list of localized texts. These texts are passed onto the LLM with a preamble indicating the task and the VQA task's question. We use the following prompts for the LLM in the question-answering experiments.

\noindent\textbf{VQAv2, OK-VQA and GQA.}

\noindent Preamble: 
\begin{lstlisting}[language=Python]
preamble = "You are a helpful assistant answering questions about images to people. You can look at the list of  object detections in the image and answer questions. The image content may not be sufficient to answer the questions, and you may need to rely on external knowledge resources or commonsense. In an image, many objects were detected. They are listed in the following format:  [object descriptions] [cx, cy, w, h], where cx is x coordinate of the center, cy is the y coordinate of the center, w is the width and h is the height of the bounding box of that object in the image."
\end{lstlisting}

\noindent Image Size:
\begin{lstlisting}[language=Python]
image_size_prompt = f"The height of the image is {image_height} and width of the image is {image_width}".
\end{lstlisting}

\noindent Image description:
\begin{lstlisting}[language=Python]
image_description = f"Full images descriptions for this image are: {image_captions}". 
\end{lstlisting}

\noindent Object representation: 
\begin{lstlisting}[language=Python]
objects_description = "The list of objects is as follows: "
for captions, (cx, cy, w, h) in zip(object_captions, object_boxes):
  objects_description += f"{captions} [{cx}, {cy}, {w}, {h}],"
\end{lstlisting}

\noindent Question prompt: 
\begin{lstlisting}[language=Python]
question_prompt = f"Q: {question} Answer in one word. A:"
\end{lstlisting}

\noindent Full prompt: 
\begin{lstlisting}[language=Python]
full_prompt = (preamble + image_size_prompt + image_description + objects_description + question_prompt)
\end{lstlisting}

\noindent \textbf{VizWiz}

\noindent Preamble: 
\begin{lstlisting}[language=Python]
preamble = "You are a helpful assistant answering questions about images to people. You can look at the list of  object detections in the image and answer questions. The image content may not be sufficient to answer the questions, and you may need to rely on external knowledge resources or commonsense. In an image, many objects were detected. They are listed in the following format:  [object descriptions] [cx, cy, w, h] [score], where cx is x coordinate of the center, cy is the y coordinate of the center, w is the width,  h is the height and score is the confidence score for the object detection. Low score means the detection is likely inaccurate, and this often makes the question unanswerable. You can answer questions as 'unanswerable'." 
\end{lstlisting}

\noindent Image Size:
\begin{lstlisting}[language=Python]
image_size_prompt = f"The height of the image is {image_height} and width of the image is {image_width}".
\end{lstlisting}

\noindent Image description:
\begin{lstlisting}[language=Python]
image_description = f"Full images descriptions for this image are: {image_captions}". 
\end{lstlisting}

\noindent Object representation: 
\begin{lstlisting}[language=Python]
objects_description = "The list of objects is as follows: "
for captions, (cx, cy, w, h) in zip(object_captions, object_boxes):
  objects_description += f"{captions} [{cx}, {cy}, {w}, {h}],"
\end{lstlisting}

\noindent Question prompt: 
\begin{lstlisting}[language=Python]
question_prompt = f"Q: {question} Answer in one word. A:"
\end{lstlisting}

\noindent Full prompt: 
\begin{lstlisting}[language=Python]
full_prompt = (preamble + image_size_prompt + image_description + objects_description + question_prompt)
\end{lstlisting}

\noindent  \textbf{MSRVTT and MSVD}

\noindent Preamble: 
\begin{lstlisting}[language=Python]
preamble = "You are a helpful assistant answering questions about videos to people. You can look at the list of  object detections in each frame and answer questions."
\end{lstlisting}

\noindent Object representation:
\begin{lstlisting}[language=Python]
objects_description = "In a video, many objects were detected in each frame."

for frame_idx in frame_idxes:
  objects_description =  f"In frame {frame_idx}, following objects were detected"
  for captions in object_captions_in_frame_idx:
    objects_description += f"{captions},"
\end{lstlisting}

\noindent Question prompt: 
\begin{lstlisting}[language=Python]
question_prompt = f"Q: {question} Answer in one word. A:"
\end{lstlisting}

\noindent Full prompt: 
\begin{lstlisting}[language=Python]
full_prompt = (preamble + objects_description + question_prompt)
\end{lstlisting}

\end{document}